\documentclass[conference]{IEEEtran}
\IEEEoverridecommandlockouts
\usepackage{cite}
\usepackage{color,xcolor}
\usepackage{multirow, multicol}
\usepackage{longtable, threeparttable, booktabs}
\usepackage{amsmath}
\usepackage{graphicx} 

\usepackage{amsmath,amsfonts}
\usepackage{algorithmic}
\usepackage{array}
\usepackage[caption=false,font=normalsize,labelfont=sf,textfont=sf]{subfig}
\usepackage{textcomp}
\usepackage{stfloats}
\usepackage{url}
\usepackage{verbatim}
\usepackage{graphicx}
\usepackage{cite}
\usepackage{textcomp}
\usepackage{xcolor}
\usepackage{balance}
\usepackage{footnote}
\usepackage{threeparttable}
\def\BibTeX{{\rm B\kern-.05em{\sc i\kern-.025em b}\kern-.08em
    T\kern-.1667em\lower.7ex\hbox{E}\kern-.125emX}}
\begin{document}

\title{Pretrain like Your Inference: Masked Tuning Improves Zero-Shot Composed Image Retrieval
\thanks{Hanjiang Lai is the corresponding author. This work is supported by the Guangdong Basic and Applied Basic Research Foundation(2024A1515012006). }
}

\author{\IEEEauthorblockN{Junyang Chen}
\IEEEauthorblockA{\textit{School of Computer Science and Engineering} \\
\textit{Sun Yat-Sen University}\\
Guangzhou, China \\
chenjy855@mail2.sysu.edu.cn}
\and
\IEEEauthorblockN{Hanjiang Lai}
\IEEEauthorblockA{\textit{School of Computer Science and Engineering} \\
\textit{Sun Yat-Sen University}\\
Guangzhou, China \\
laihanj3@mail.sysu.edu.cn}
}

\maketitle

\begin{abstract}
  Zero-shot composed image retrieval (ZS-CIR), which takes a textual modification and a reference image as a query to retrieve a target image  without triplet labeling, has gained more and more attention in data mining. Current ZS-CIR research mainly relies on the generalization ability of pre-trained vision-language models, e.g., CLIP. However, the pre-trained vision-language models and CIR tasks have substantial discrepancies, where the vision-language models focus on learning the similarities but CIR aims to learn the modifications of the image guided by text. In this paper, we introduce a novel unlabeled and pre-trained masked tuning approach, which reduces the gap between the pre-trained vision-language  model and the downstream CIR task. First, to reduce the gap, we reformulate the contrastive learning of the vision-language model as the CIR task, where we randomly mask input image patches to generate $\langle$masked image, text, image$\rangle$ triplet from an image-text pair. Then, we propose a simple but novel pre-trained masked tuning method, which uses the text and the masked image to learn the modifications of the original image. With such a simple design, the proposed masked tuning can learn to better capture fine-grained text-guided modifications. Extensive experimental results demonstrate the significant superiority of our approach over the baseline models on four ZS-CIR datasets, including FashionIQ, CIRR, CIRCO, and GeneCIS. Our codes are available at \url{https://github.com/Chen-Junyang-cn/PLI}
\end{abstract}


\section{Introduction}

\begin{figure}[t]
    \centering
    \includegraphics[width=1\linewidth]{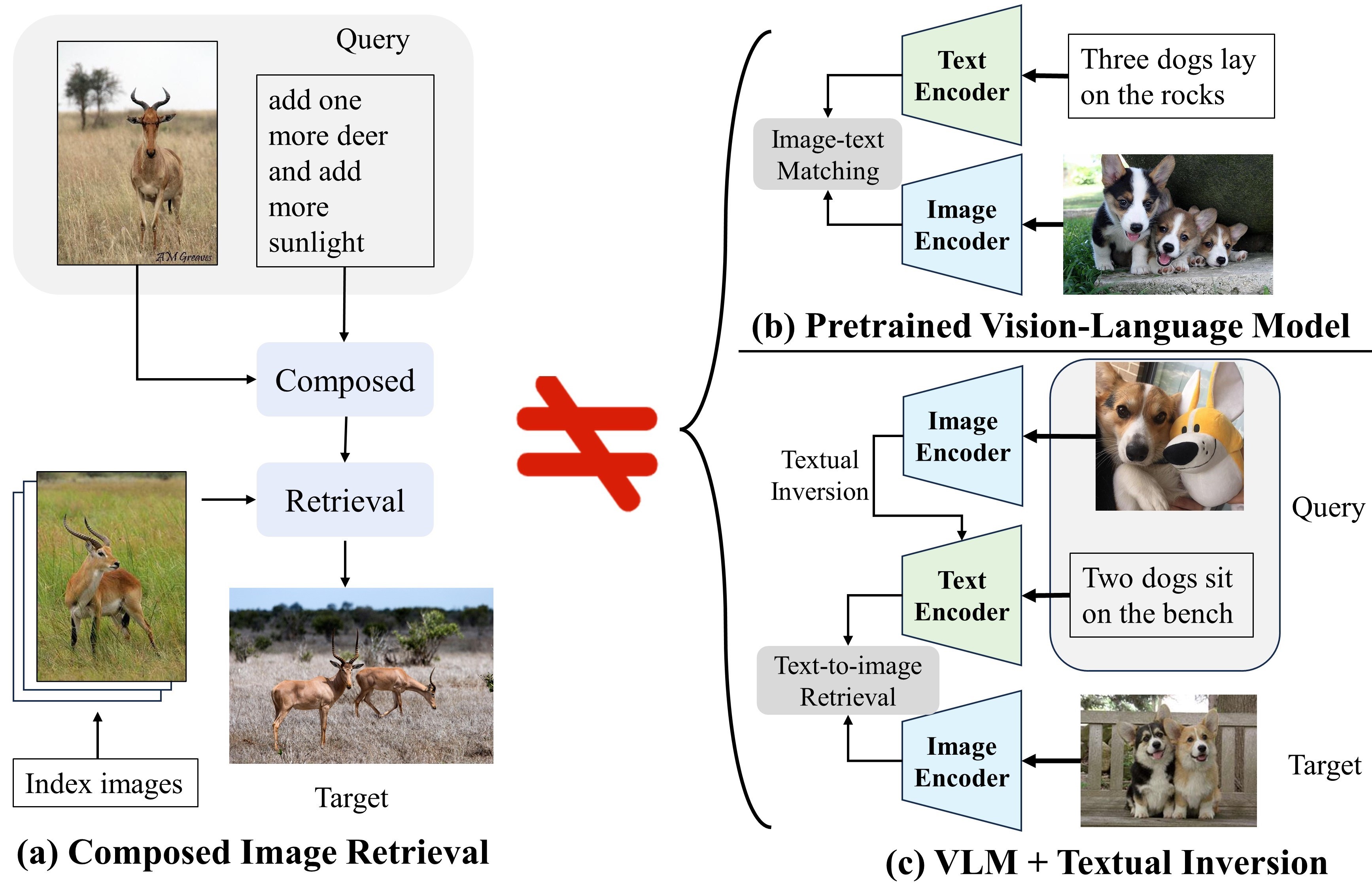}
    \caption{(a) Workflow of composed image retrieval task, which is essentially different from the pre-trained VLM objective and textual inversion. (b) The pre-trained vision-language models~\cite{li2022blip, radford2021clip} are to align text and image features. 
    (c) The recent ZS-CIR methods~\cite{Baldrati_2023_ICCV, saito2023pic2word} also introduced textual inversion into the pre-trained VLM, which mapped the reference image into the text domain, to further improve the performance.}
    \label{fig:intro}
\end{figure}

\IEEEPARstart{R}{ecently}, composed image retrieval (CIR)~\cite{baldrati2022effective, liu2021image,lee2021cosmo, xu2024csvt} has gained more and more attention, where it allows users to specify textual modifications to a reference image to retrieve a relevant image. 
This multi-modal input enables users to conduct fine-grained retrieval, as the retrieved data must not only contain user-specified modifications but also remain consistent with the reference image~\cite{vo2019composing}, as shown in Figure~\ref{fig:intro} (a). Hence, CIR has garnered significant attention due to its potential applications in various domains, including online retail and Internet search~\cite{jandial2022sac, wu2021fashion}.

Many supervised methods~\cite{baldrati2022effective, baldrati2022conditioned, liu2023bi} have been proposed for CIR.  Current advanced work on CIR relies on the impressive capabilities of the vision-language models (VLMs)~\cite{li2022blip,radford2021clip}  pre-trained on large-scale datasets with image-text pairs. Thus these supervised CIR methods finetune the VLMs with the labelled triplets, and an extra fusion network that combines the textual modification and the reference image has also been proposed. For instance, Baldrati \textit{et al.} ~\cite{baldrati2022conditioned} employed a two-stage methodology, where the CLIP~\cite{radford2021clip} text encoder was first fine-tuned and a fusion network was then trained to combine the reference image and text features. Liu \textit{et al.}~\cite{liu2023bi} took a more advanced step by employing a bidirectional training strategy to fine-tune the BLIP~\cite{li2022blip} and the fusion network. In summary, these supervised methods require a large number of labeled triplets $\langle$reference image, text, target image$\rangle$ to support fully supervised training. Nevertheless, labeling these triplets is a costly endeavor. 

To reduce the human-annotated triplets, a new task, called zero-shot CIR (ZS-CIR)~\cite{Baldrati_2023_ICCV, saito2023pic2word}, is proposed to train the retrieval models without the costly manual labeled data. Without the supervised information, recent ZS-CIR approaches will be more dependent on the generalization ability of the pre-trained vision-language model. 
Karthik \textit{et al.} \cite{karthik2024visionbylanguage} captioned the reference image by using large language model (LLM) 
for ZS-CIR. 
Li \textit{et al.} \cite{liimproving} leveraged LLM to enhance the web-collected image-text pairs. And then they further improved the alignment of visual and linguistic input using the generated dataset.

However, due to the gap between the CIR task and the pre-trained VLM, performance may not be satisfied by simply applying VLMs to ZS-CIR. To be specific, as illustrated in Figure~\ref{fig:intro} (b), the objective of the pre-trained models is to align features from image and text modalities. For example, the image of a white cloth should be similar to the word ``white cloth'' (VLMs). While the CIR task aims to learn the modification of the reference image guided by text. Given the image of a white cloth and the text ``change it to black cloth'', the goal is not to learn the semantic similarity  between the image and text, but to modify the image according to the text.

To reduce the gap, some approaches \cite{Baldrati_2023_ICCV, saito2023pic2word, cohen2022my,  gu2023languageonly} have proposed to train an additional module for VLMs. These methods froze parameters of the VLM, and the additional module, e.g., textual inversion~\cite{Baldrati_2023_ICCV}, was trained to improve performance. As shown in Figure~\ref{fig:intro} (c) in~\cite{Baldrati_2023_ICCV} and \cite{saito2023pic2word}, the textual inversion/pic2word was first trained to translate the reference image into a pseudo-word vector. Then, the pseudo-word vector was concatenated with the textual feature of the modification to obtain the query feature. Finally, the target image was retrieved by utilizing the efficient alignment capabilities of VLM. In recent concurrent work, LinCIR \cite{gu2023languageonly} employed a random noise addition strategy to train inversion networks on text-only data.

In this paper, we address this problem in a different way: we directly translate the downstream CIR task into the pre-trained VLM task without any additional module. A novel and straightforward self-supervised pre-trained masked tuning approach is proposed to bridge the gap between  VLM and CIR tasks. The pivotal insight lies in adapting the pre-trained vision-language task to be more similar to the CIR task. To be more precise, we first make the pre-trained image-text pair data in the same form as the CIR's data, i.e., changing the image-text pair $\langle T, I \rangle$  to the triplet form  $\langle I_m, T, I \rangle$. This triplet comprises a masked image, text, and the original image, where the masked image is obtained by randomly masking the original image. Then, we use the text $T$ to guide the masked image $I_m$ to obtain the fused feature, which is close to the feature of the original image $I$. As an illustration, consider the scenario where ``a dog'' and ``bench'' are masked in the masked image $I_m$ accompanied by the text ``two dogs sit on the bench''. In this context, the text signifies the necessary modifications to the masked image $I_m$ (another dog and bench should be added), and the masked image also ensures that the target image $I$ should be consistent with the masked image in unspecified attributes, like the color and breed of the unmasked dog. This aligns perfectly with the requirements of the CIR task. 

While the method is straightforward, the gains are amazing, offering an intriguing avenue for further exploration of ZS-CIR. Experiments demonstrate notable improvements in our approach compared to state-of-the-art baselines on four benchmark datasets: FashionIQ, CIRR, CIRCO, and GeneCIS. Particularly, on the FashionIQ dataset with Recall@10 and Recall@50, our method shows remarkable increases of up to 9.75\% and 12.46\%, respectively. Moreover, our proposed method also outperforms previous state-of-the-art ZS-CIR models in the case of ``frozen VLM +  additional module'', which demonstrates the high scalability and efficiency of our approach.

Our contributions can be summarized as follows:
\begin{itemize}
    \item We propose a novel pre-trained vision-language model that visually bridges the gap between pre-trained models and CIR tasks by simply employing masked tuning.
    \item Our proposed method enriches the CIR triplets by masking image-text pairs. This self-supervised pre-training technology ultimately reduces the retrieval model's reliance on costly triplet annotation data.
    \item Significant performance enhancements are observed in ZS-CIR on four benchmark datasets: FashionIQ, CIRR, CIRCO, and GeneCIS. Our proposed method surpasses the baselines, establishing itself as the SOTA model.
\end{itemize}

\section{Related Work}
\subsection{Composed Image Retrieval}
Multi-modal compositional learning has gained widespread attention in diverse data mining tasks \cite{meng2024csvt, wei2023csvt}, including visual question answering~\cite{li2021align, wu2024tmmResolving, chen2024csvt_vqa}, image captioning \cite{hu2022scaling, yu2020csvt_ic}, and image generation~\cite{rombach2022high, gao2024csvt_ig}. The composed image retrieval (CIR) pertains to multi-modal compositional learning, which focuses on retrieving a target image through the utilization of joint embedding features derived from compositions of a reference image and text~\cite{vo2019composing}. The field of CIR has seen significant exploration in the domain of fashion~\cite{wu2021fashion}, and more recently, it has been extended to real-life images~\cite{liu2021image}. 

Contemporary mainstream CIR research employs a supervised learning paradigm, wherein combinatorial multimodal information is acquired when training a CIR model. Cutting-edge CIR models~\cite{baldrati2022effective, liu2021image, lee2021cosmo, liu2023bi} hinged on a post-fusion strategy, i.e.,  fusion occurs subsequent to extracting visual and linguistic features using potent pre-trained encoders like CLIP and BLIP. Liu \textit{et al.}~\cite{liu2021image} demonstrated strong performance through the fine-tuning of CLIP text encoders for CIR tasks, along with a combiner network to fuse image-text CLIP features. Further, Liu \textit{et al.}~\cite{liu2023bi} improved performances via the bidirectional training scheme on the combiner and the BLIP-based pre-trained model. Another line of research focuses on multi-granularity matching instead of network structure.
Chen \textit{et al.}~\cite{chen2022composed} simulated multi-granularity queries by introducing normally distributed noise into the feature space. Furthermore, Chen and Lai~\cite{chen2023ranking} observed that current triple optimization methods tend to overlook semantic diversity and fail to model many-to-many mapping relationships with the necessary level of uncertainty. 
Limited by collecting triple data for CIR requires expensive human resources, the existing CIR training data such as FashionIQ \cite{wu2021fashion} and CIRR \cite{liu2021image} are usually small.
Some approaches \cite{liu2023zero, gu2023compodiff, vaze2023genecis, LaSCo_2024, Jang_2024_CVPR} have investigated automatic methods to generate CIR triplets to scale up the dataset sizes.
Liu \textit{et al.} \cite{liu2023zero} automatically constructed triplets for training CIR models by leveraging the large-scale image caption data Laion-COCO.
Gu \textit{et al.} employed a text-conditional diffusion model and CLIP-based filtering to automatically obtain high-quality and large-scale CIR triplets dataset SynthTriplets18M \cite{gu2023compodiff}.
Vaze \textit{et al.} leveraged an off-the-shelf text-toscene-graph parser to extract relationships within captions to mine conditional similarity triplet training data.
LaSCo \cite{LaSCo_2024} dataset was created with minimal manual effort from the VQA2.0 data. LaSCo generated triplets by pairing images with complementary images and converting the question-answer pairs into modified text by GPT-3 \cite{brown2020language}.
CoVR \cite{CoVR_2024} generated triplet data using fine-tuned LLM to produce visual differences from reference and target captions.
VDG \cite{Jang_2024_CVPR} generated triplets by fine-tuning the large language model (LLM) to accurately describe the difference between the reference and target images.

However, all of these methods require a lot of training triplets on the CIR datasets. Hence, recent advancements in CIR research have delved into the utilization of unlabeled pre-trained models to adapt to the downstream CIR task, notably in the form of zero-shot CIR (ZS-CIR). 
Pic2Word~\cite{saito2023pic2word} employed the pre-trained vision-language model in addition to unlabeled image datasets to train a textual inversion network. This network was designed to convert input images into linguistic tokens, enabling the flexible combination of image and text queries. SEARLE~\cite{Baldrati_2023_ICCV} employed image data and GPT-powered regularization to train a textual inversion network to generate a set of pseudo-word tokens. 
Recent studies \cite{karthik2024visionbylanguage, liimproving} have explored the use of LLM for the ZS-CIR task. 
Fromage \cite{koh2023grounding} directly mapped between image and text embedding spaces by fine-tuning the input and output linear layers for cross-modal compositional retrieval.
CIReVL~\cite{karthik2024visionbylanguage} combined large-scale VLM and LLM to transform CIR into text retrieval task. MCL \cite{liimproving} first generated large-scale multimodal compositional datasets from LLM. The generated dataset was then used to fine-tune the LLM to facilitate the alignment of visual features and linguistic space.
In concurrent work, Context-I2W \cite{tang2023contexti2w} proposed a context-dependent mapping network to dynamically map a description-relevant image to a word. LinCIR~\cite{gu2023languageonly} trained a textual inversion network with text datasets by adding noise and self-masking projection. Du \textit{et al.} \cite{du2024imagesentence} proposed a new textual inversion network that maps an image to a sentence in an asymmetric retrieval context.
In contrast to these textual inversion-based approaches, our methodology adopts a novel pre-training paradigm. We choose to simply fine-tune the vision-language contrastive learning model, which can reduce the gap between the pre-training model and the CIR task. Furthermore, our approach eliminates the necessity for additional unlabeled image datasets typically required in the training of textual inversion models.

\subsection{Masked Modeling}
Masked language modeling~\cite{brown2020language, kenton2019bert, zhang2024tmmmask} stands as a highly successful unsupervised representation learning approach in natural language processing (NLP). It involves the retention of a portion of the input sequence, and the pre-training model is trained to predict the missing content. This mask-based pre-training paradigm exhibits the robust ability to generalize to a variety of downstream NLP tasks, showcasing strong scalability~\cite{brown2020language}.
Inspired by the accomplishments in NLP, various image-specific masking techniques have been proposed \cite{dosovitskiy2020vit, wen2024image, fang2024mae}. ViT~\cite{dosovitskiy2020vit} directly applied a standard Transformer~\cite{vaswani2017attention} to an image and predicted masked patches. In contrast, iGPT~\cite{chen2020generative} and BEiT~\cite{bao2021beit} respectively make predictions for masked pixels and discrete tokens. Notably, MAE~\cite{he2022masked} capitalized on the advantages of masking and exclusively inputs visible blocks into ViT, resulting in improved training speed.

Most recently, FLIP \cite{li2023scaling} has advanced masked modeling in vision-language models. 
FLIP is the most similar work to us. While FLIP \cite{li2023scaling}  emphasized the scaling of sparse computing implementation by masking, involving a trade-off between the benefits of processing more sample pairs and the potential degradation of sample-by-sample encoding. Although our proposed method also employed a masking model, our primary motivation lies in exploring how masking can serve as a bridge between pre-trained models and their application in downstream CIR tasks. Also, the inputs of FLIP were image-text pairs while our method took a triplet as input. The goals are totally different.

\begin{figure*}[!t]
    \centering
    \includegraphics[width=0.99\linewidth]{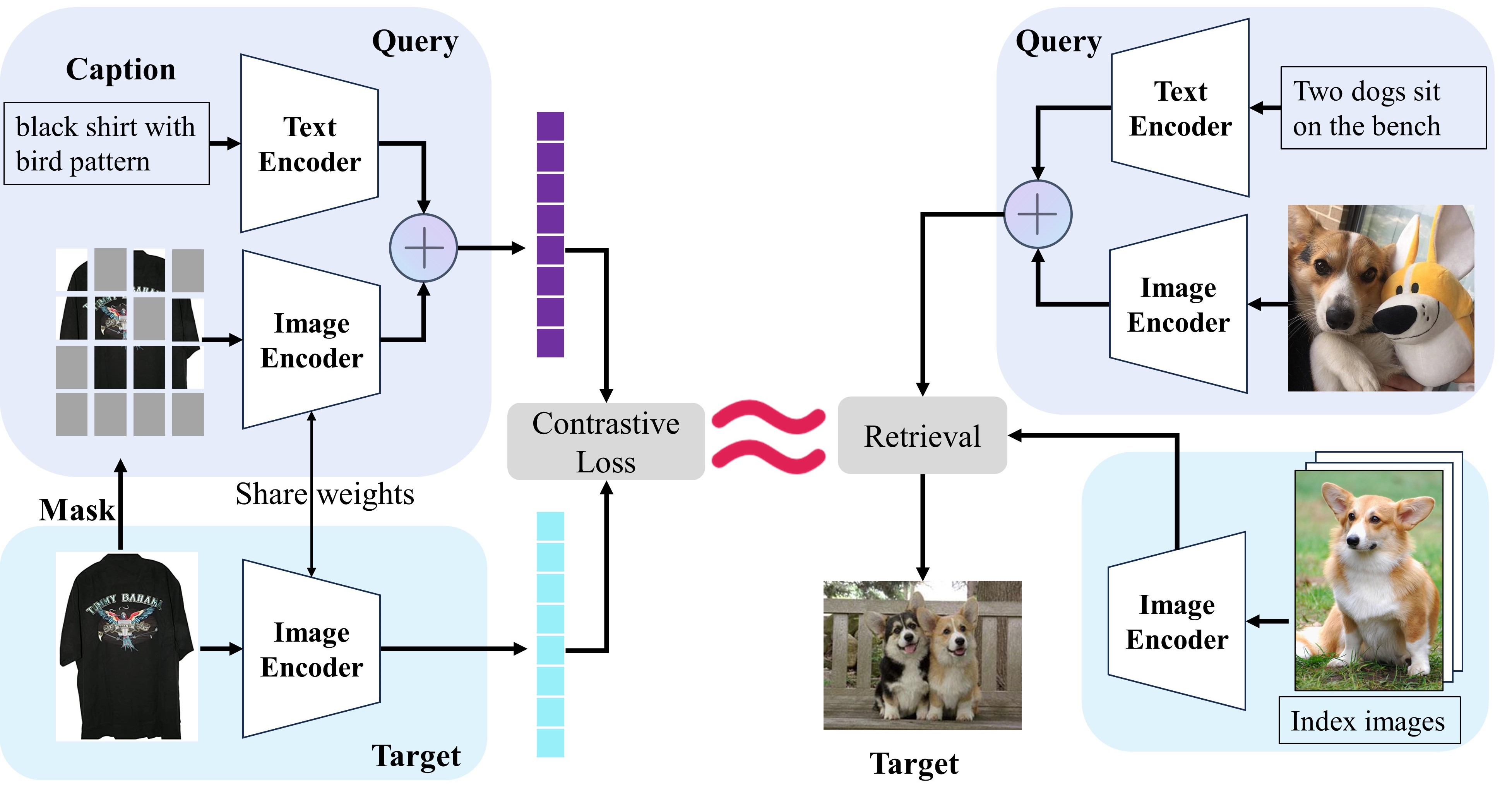}
    \caption{Overview of our masked pre-training method. \textit{Left}: we randomly apply a high masking ratio to mask image patches, and let the pre-trained task approximate the CIR task. \textit{Right}: we leverage the pre-trained model  at inference time on ZS-CIR.}
    \label{fig:overview}
\end{figure*}

\section{Method}

\subsection{Problem Definition}
In the settings of zero-shot composed image retrieval (ZS-CIR), we do not have any training data from the CIR dataset. Please note that not only the supervised triplet data but also unsupervised data is not available.  ZS-CIR mainly builds upon the pre-trained vision-language models, e.g., CLIP. Similar to the previous works~\cite{saito2023pic2word,Baldrati_2023_ICCV}, 
in this paper, we only use the image-text pairs $\langle T, I \rangle$, which are also used in the pre-trained vision-language model, to further reduce the gap.

\subsection{Masked Tuning}
As illustrated in Figure~\ref{fig:overview}, given the image-text pairs $\langle T, I \rangle$, our method utilizes a masking approach to generate an approximate triplet for CIR to adjust the pre-trained visual-language model. 
Specifically, we employ a masking strategy to deliberately omit image semantics and obtain the masked image $I^m$. Meanwhile, the original caption $T$ can be naturally interpreted as a modification applied to the masked image. With our masking strategy, we efficiently create a text-to-image modification query for the input. And the unmasked original image serves as ``task supervision''. Our approach extends from image-text pairs to CIR triplets $\langle I^m, T, I \rangle$, which effectively reduces the gap between pre-trained models and CIR tasks. The key components of the proposed approach are described below.

Initially, we partition the image into non-overlapping blocks of a fixed size. The vision transformer (ViT) \cite{dosovitskiy2020vit} is utilized as the image encoder. Next, we randomly mask the blocks according to a uniform distribution and then utilize the remaining visible blocks as input to the ViT. We apply a high masking rate (50\% or 75\%) to obscure a majority of the blocks. This choice is deliberate and aimed at preventing the pre-trained model from depending on the extrapolation of semantic information from nearby visible blocks. Considering the inherent redundancy in image information \cite{he2022masked, li2023scaling}, the elevated masking rate compels the model to amalgamate the semantic information from the unmasked blocks with textual input to ``reconstruct'' the original unmasked image semantics.


The objective of the pre-training model is to minimize the cosine similarity between the combined query features and the target features, achieved through contrastive learning. Considering a multimodal dataset with image-text pairs, denoted as $S = \{(T_n, I_n)\}^N_{n=1}$, where $I$ and $T $ represent images and texts, respectively. By masking most of the blocks in the image $I$, we obtain the masked image $I^m$. Consequently, we have a dataset $S^m = \{(I_n^m, T_n, I_n)\}^{N}_{n=1}$, where $I_n^m$ and $T_n$ serve as query inputs and $I_n$ is the target image. As shown in Figure \ref{fig:overview}, the masked image and text of the query are separately encoded by the image encoder and text encoder, after which the combined feature $f^q$ is derived by simply adding the two features. Simultaneously, the target image $I_n$ is fed through the image encoder, which employs shared weights, resulting in the target feature $f^t$. Subsequently, we train the image/text encoder to minimize the contrastive loss, utilizing other samples from the same batch as negative pairs to perform contrastive learning~\cite{chen2020simple}:
\begin{equation}
    \mathcal{L}_{CL}\left(f^q,f^t\right)=\frac1B\sum_{i=1}^B-\log\frac{\exp\left(\kappa\left(f^q_i,f^t_i\right)\right)}{\sum_{j=1}^B\exp\left(\kappa\left(f^q_i,f^t_j\right)\right)},
\end{equation} where $B$ is mini-batch size and $\kappa(\cdot)$ is cosine similarity function. It's noteworthy that, in accordance with the findings from research \cite{liu2023bi}, $f^q$ and $f^t$ undergo no additional projection or normalization during the pre-training phase. 
This contrastive loss compels the output of the combined feature to closely resemble the feature of the target image.

\subsection{Inference}
Thanks to the alignment between the motivation for pre-training and the requirements of the CIR task, we can effortlessly employ the pre-trained model to retrieve the target image, as depicted in Figure \ref{fig:overview}. During the inference stage, the user provides the input $\langle I_r, T_r \rangle$ for the query. Subsequently, we obtain the combined feature $f_r$ by employing both the image encoder and text encoder as $f_r = (1-w)f_r^I + f_r^T$, where $w$ represents the mask ratio and $f_r^I$ and $f_r^T$ denote the image and text features. Note that the hyperparameter $1-w$ is fixed in the test phase when the value of masking ratio $w$ is determined in the pre-training phase.
We use the weighting method since we use the masked image as input in the pre-training model while we utilize the whole image as input in testing, which may lead to a distribution shift as discussed in FLIP~\cite{li2023scaling}. Hence, we propose the weighted method to alleviate this shift. Next, the features of the image dataset are also processed using the same image encoder. We locate the candidate image $I_t$ by comparing it with the combined feature. 

\textit{Remark:} We emphasize heavily that the novelty of this work lies in proposing a simple and effective approach for adapting the pre-trained model to the CIR task. Existing ZS-CIR methods based on textual inversion \cite{Baldrati_2023_ICCV, saito2023pic2word, gu2023languageonly} exploit the alignment capability of the VLM and ignore the gap between the pre-trained VLM and the CIR task. These methods aim to reduce CIR to standard text-to-image retrieval.
In contrast, our approach directly reduces the gap between the VLM and the CIR task by creating a text-to-image modification query. 
Consequently, our approach makes the pre-training process more similar to an inference process than existing methods, which provides a novel line of research for solving the ZS-CIR task.

\section{Experiments}

\subsection{Experimental Settings}
\textbf{Pre-training dataset:} We use two datasets with a total of 695K image-text pairs to pre-train our ZS-CIR model:  
\begin{itemize}
    \item LLaVA-CC3M-Pretrain-595K\footnote{https://huggingface.co/datasets/liuhaotian/LLaVA-CC3M-Pretrain-595K}: This dataset contains 595K image-text pairs from the CC3M dataset \cite{sharma2018conceptual}. It is more balanced filtered and used to align textual and modal features of LLaVA \cite{liu2024visual}.
    \item 
    We follow \cite{Baldrati_2023_ICCV} and use the unlabeled test split of ImageNet1K \cite{russakovsky2015imagenet} as the pre-training dataset.  We employ the BLIP2-FlanT5-XL\footnote{\url{https://huggingface.co/Salesforce/blip2-flan-t5-xl}} vision-language model to generate a corresponding caption for each image. We finally obtained 100K image-text pairs.
\end{itemize}
Please note that our pre-training data comprises significantly fewer data compared to the Pic2Word \cite{saito2023pic2word} and LinCIR \cite{gu2023languageonly}, accounting for approximately 23\% and 13\%, respectively. 

We follow the experimental settings of \cite{baldrati2022conditioned} to utilize the AdamW optimizer~\cite{loshchilov2019decoupled} with a learning rate $10^{-6}$ and weight decay $5 \times 10^{-5}$. Contrastive learning is conducted with a batch size of 64. 
We evaluate the proposed method by fine-tuning the pre-trained models. We choose two different pre-trained models CLIP ViT-B/32 and CLIP ViT-L/14. 
The mask ratios of pre-trained CLIP models are set to 75\%. CLIP ViT-B/32 
is trained with one NVIDIA RTX3090, and CLIP ViT-L/14 is trained with four NVIDIA RTX3090. {The time to train CLIP ViT-L on four RTX3090 is about 0.036 seconds per image with batch size 64.} In the following table of results, we refer to ViT-B/32 and ViT-L/14 as B/32 and L/14, respectively.  

All proposed methods are implemented using the PyTorch \cite{NEURIPS2019_bdbca288} framework. To promote transparency and reproducibility, we will release the entire experimental code repository as open-source.

\subsection{Datasets and Baselines}
We conduct and evaluate our method on four datasets: FashionIQ \cite{wu2021fashion}, CIRR \cite{liu2021image}, CIRCO \cite{Baldrati_2023_ICCV}, and GeneCIS \cite{vaze2023genecis}. We follow the standard evaluation protocols \cite{baldrati2022effective, liu2021image, Baldrati_2023_ICCV,  vaze2023genecis} to test our method.

FashionIQ \cite{wu2021fashion} is a fashion dataset and contains three distinct product categories: dresses, shirts, and toptee. In line with prior studies \cite{lee2021cosmo}, we adopt Recall@10 (R@10) and Recall@50 (R@50) as evaluation metrics and calculate the average performance across these three subsets to gauge overall effectiveness. Following the previous practice \cite{Baldrati_2023_ICCV, saito2023pic2word, Chen_2020_CVPR}, all results are evaluated on the validation set since the test set is unavailable.

Compose Image Retrieval on Real-life Images (CIRR) \cite{liu2021image} dataset consists of over 36,000 image-text pairs. The images in this dataset are from NLVR$^2$ \cite{suhr-etal-2019-corpus} and are divided into multiple subsets of six images. The images in the subset are semantically and visually similar. This dataset is curated to address the limitations found in existing datasets, notably the need for greater visual complexity and the prevalence of false negatives. 
Following the evaluation protocols~\cite{liu2021image} in previous CIRR studies, we employ Recall@K (where K = 1, 5, 10, 50) and Recall$_{\text{Subset}}@K$ (where K = 1, 2, 3) as the evaluation metrics. Notably, Recall$_{\text{Subset}}@K$ exclusively considers candidate pairs within the same subset.

The Composed Image Retrieval on Common Objects in context (CIRCO) \cite{Baldrati_2023_ICCV} dataset 
comprises a total of 1,020 queries, and each query has an average of 4.53 ground truth values. The number of queries in the CIRCO validation set and test set is 220 and 800, respectively. Notably, CIRCO uses entirety of 120K images of COCO \cite{lin2014microsoft} as an index set, which introduces more  distractors.
Since CIRCO has multiple ground truths, we follow the previous evaluation criterion~\cite{Baldrati_2023_ICCV} and use mean Average Precision (mAP) as a more fine-grained metric. Specifically, we used mAP@K with $K= 5, 10, 25, 50$.

General Conditional Image Similarity (GeneCIS) \cite{vaze2023genecis} is a challenging dataset to adapt to a range of similarity conditions. 
There are four conditional evaluation tasks in GeneCIS: (1) focus on an attribute, (2) change an attribute, (3) focus on an object, and (4) change an object. 
VAW dataset \cite{pham2021learning} is used to construct the two tasks for the attributes of `focus on' and `change'. Two tasks for `focus on' and `change' objects are based on the COCO Panoptic Segmentation data \cite{lin2014microsoft}. 
Each task has about 2,000 queries that require the model to retrieve a unique target image from a gallery of similar images. The gallery size ranges from 10 to 15, and it is guaranteed that there is only one positive image among the gallery images.

We compare several zero-shot baseline models for a fair evaluation of our approach.
\begin{itemize}
    \item Baseline (Image + Text): This baseline method retrieves the target images by adding the reference image's feature with the feature of the corresponding caption.
    \item Text-only: Similarity is assessed solely based on the CLIP text features of the captions.
    \item Image-only: This method retrieves the target images based on the reference images.
    \item Captioning~\cite{Baldrati_2023_ICCV}:  The caption generated from the reference image is used for ZS-CIR.
    \item PALAVRA~\cite{cohen2022my}: A two-stage approach is employed for open-world retrieval: 1) learning an inversion mapping that maps CLIP image space to its word embedding, and 2) learning the word embedding of a new personalized concept.
    \item Pic2Word~\cite{saito2023pic2word}: It leverages the pre-trained model using the textual inversion network.
    \item SEARLE~\cite{Baldrati_2023_ICCV}: It maps the visual feature space of the reference image to  the CLIP text embedding space, which is also integrated with the GPT-powered regularization loss~\cite{brown2020language}.
    \item  LinCIR~\cite{gu2023languageonly}: It only uses language for CIR training, which employs a self-masking projection to train the text inversion network on the 5.5M training captions.
\end{itemize}

\subsection{Main Results}

\begin{table*}[!ht]
\centering
\caption{Quantitative Evaluation Results on FashionIQ Validation Set. The Best and Second-Best Scores Are Highlighted in Bold and Underlined, Respectively. $^{\dagger}$ and $^\S$ Indicates Results Are Cited \cite{Baldrati_2023_ICCV} and \cite{saito2023pic2word}, Respectively.}
  \begin{tabular}{c | l | c c | cc | cc | cc} 
  \hline
  \multicolumn{1}{c|}{} & \multicolumn{1}{c|}{} & \multicolumn{2}{c|}{Shirt} & \multicolumn{2}{c|}{Dress} &\multicolumn{2}{c|}{Toptee} &\multicolumn{2}{c}{Average} \\
  \cline{3-4}
  \cline{5-6}
  \cline{7-8}
  \cline{9-10}
  \multicolumn{1}{c|}{Backbone} & \multicolumn{1}{l|}{Method} & R$@10$ & R$@50$ & R$@10$ & R$@50$ & R$@10$ & R$@50$ & R$@10$ & R$@50$ \\ 
  \hline
  \multirow{8}{*}{CLIP B/32} & Image-only$^{\dagger}$ & 6.92 & 14.23 & 4.46 & 12.19 & 6.32 & 13.77 & 5.90 & 13.37 \\
  & Text-only$^{\dagger}$ & 19.87 & 34.99 & 15.42 & 35.05 & 20.81 & 40.49 & 18.70 & 36.84 \\ 
  & Captioning$^{\dagger}$~\cite{Baldrati_2023_ICCV} & 17.47 & 30.96 & 9.02 & 23.65 & 15.45 & 31.26 & 13.98 & 28.62 \\
  & PALAVRA$^{\dagger}$ \cite{cohen2022my} & 21.49 & 37.05 & 17.25 & 35.94 & 20.55 & 38.76 & 19.76 & 37.25 \\
  & SEARLE-OTI$^{\dagger}$~\cite{Baldrati_2023_ICCV} & \underline{25.37} & {41.32} & {17.85} & \underline{39.91} & {24.12} & {45.79} & {22.44} & {42.34} \\
  & SEARLE$^{\dagger}$~\cite{Baldrati_2023_ICCV} & {24.44} & \underline{41.61} & \underline{18.54} & {39.51} & \underline{25.70} & \underline{46.46} & \underline{22.89} & \underline{42.53}  \\ 
  & Baseline$^{\dagger}$ & 13.83 & 30.88 & 13.44 & 26.25 & 17.08 & 31.67 & 14.78 & 29.60 \\
  & Ours & \textbf{32.72} & \textbf{51.33} & \textbf{26.89} & \textbf{48.98} & \textbf{33.27} & \textbf{56.71} & \textbf{30.96} & \textbf{52.34}\\
  \hline
  \multirow{6}{*}{CLIP L/14} & Pic2Word$^{\S}$ \cite{saito2023pic2word} & 26.20 & 43.60 & 20.00 & 40.20 & 27.90 & 47.40 & 24.70 & 43.70 \\
  & {SEARLE-XL-OTI}$^{\dagger}$~\cite{Baldrati_2023_ICCV} & \underline{{{30.37}}} & \underline{{{47.49}}} & \underline{{{21.57}}} & \underline{{{44.47}}} & \underline{{{30.90}}} & \underline{{{51.76}}} & \underline{{{27.61}}} & \underline{{{47.90}}} \\
  & {SEARLE-XL}$^{\dagger}$~\cite{Baldrati_2023_ICCV} & {{26.89}} & {{45.58}} & {{20.48}} & {{43.13}} & {{29.32}} & {{49.97}} & {{25.56}} & {{46.23}} \\ 
  & LinCIR \cite{gu2023languageonly} & 29.10 & 46.81 & 20.92 & 42.44 & 28.81 & 50.18 & 26.28 & 46.49 \\
  & Baseline$^{\S}$ & 21.00 & 34.50 & 16.30 & 33.60 & 22.20 & 39.00 & 19.80 & 35.70 \\
  & Ours & \textbf{39.02} & \textbf{58.85} & \textbf{29.45} & \textbf{53.26} & \textbf{40.65} & \textbf{64.22} & \textbf{36.37} & \textbf{58.78} \\
  \hline
  \end{tabular}
  
  \label{tab:fashioniq_val}
\end{table*}

\begin{table*}[!ht]
 \centering
  \caption{Quantitative Results on CIRR Test Set. Best and Second-Best Scores Are Highlighted in Bold and Underlined, Respectively. $^{\dagger}$ and $^\S$ Indicates Results Are Cited~\cite{Baldrati_2023_ICCV} and \cite{saito2023pic2word}, Respectively. -- Denotes Results Not Reported in the Original Paper.}
  \begin{tabular}{c|l|cccc|ccc} 
  \hline
  \multicolumn{1}{c|}{} & \multicolumn{1}{c|}{} & \multicolumn{4}{c|}{Recall$@K$} & \multicolumn{3}{c}{Recall$_{\text{Subset}}@K$} \\
  \cline{3-6}
  \cline{7-9}
  \multicolumn{1}{l|}{Backbone} & \multicolumn{1}{l|}{Method} & $K = 1$ & $K = 5$ & $K = 10$ & $K = 50$ & $K = 1$ & $K = 2$ & $K = 3$ \\ 
  \hline
  \multirow{8}{*}{CLIP B/32} & Image-only$^{\dagger}$ & 6.89  & 22.99  & 33.68 & 59.23 & 21.04 & 41.04 & 60.31 \\ 
  & Text-only$^{\dagger}$ & 21.81 & 45.22 & 57.42 & 81.01 & \textbf{62.24} & \textbf{81.13} & \textbf{90.70} \\ 
  
  & Captioning$^{\dagger}$ & 12.46 & 35.04 & 47.71 & 77.35 & 42.94  & 65.49 & 80.36 \\
  & PALAVRA$^{\dagger}$ \cite{cohen2022my} & 16.62 & 43.49 & 58.51 & 83.95 & 41.61 & 65.30 & 80.94 \\
  & {SEARLE-OTI}$^{\dagger}$~\cite{Baldrati_2023_ICCV} & \underline{24.27} & {53.25} & {66.10} & {88.84} & 54.10 & 75.81 & 87.33 \\
  & {SEARLE}$^{\dagger}$~\cite{Baldrati_2023_ICCV} & {24.00}  & \underline{53.42} & \underline{66.82} & \underline{89.78} & {54.89} & {76.60} & {88.19}  \\
  & Baseline$^{\dagger}$ & 11.71 & 35.06 & 48.94 & 77.49 & 32.77  & 56.89 & 74.96 \\
  & Ours & \textbf{24.31} & \textbf{53.76} & \textbf{67.78} & \textbf{90.07} & \underline{55.66} & \underline{77.57} & \underline{89.59} \\
  \hline
  
  \multirow{6}{*}{CLIP L/14} & Pic2Word$^\S$ \cite{saito2023pic2word} & 23.90 & 51.70 & 65.30 & 87.80 & -- & -- & -- \\
  & {SEARLE-XL-OTI}$^{\dagger}$~\cite{Baldrati_2023_ICCV} & {24.87} & {52.31} & {66.29} & {88.58} & {53.80} & {74.31} & {86.94}  \\
  & {SEARLE-XL}$^{\dagger}$~\cite{Baldrati_2023_ICCV} 
  & {{{24.24}}} & {{{52.48}}} & {{{66.29}}} & \underline{{88.84}} & {{53.76}} & {{75.01}} & {{88.19}} \\
  & LinCIR \cite{gu2023languageonly} & \underline{25.04} & \underline{53.25} & \underline{66.68} & -- & \textbf{57.11} & \underline{77.37} & \underline{88.89} \\
  & Baseline$^\S$ & 12.40 & 36.20 & 49.10 & 78.20 &--&--&--\\
  & Ours & \textbf{26.15} & \textbf{56.82} & \textbf{69.30} & \textbf{89.76} & \underline{56.22} & \textbf{77.52} & \textbf{89.74} \\
  
  
  \hline
  \end{tabular}
  \label{tab:cirr_test}
\end{table*}

\subsubsection{Quantitative Results on FashionIQ}

Table~\ref{tab:fashioniq_val} shows the comparison results of the proposed method and the state-of-the-art baselines on the FashionIQ validation dataset. In general, our method demonstrates a substantial performance gain over the baselines across all three categories. This substantial improvement over the state-of-the-art methods underscores the  effectiveness of our approach in the FashionIQ validation set.

With the CLIP ViT-B/32 backbone network, SEARLE \cite{Baldrati_2023_ICCV} is the previous state-of-the-art (SOTA) method. Notably, our method exhibits a significant improvement in both the Recall@10 and Recall@50 metrics compared to SEARLE. To be precise, our method outperforms the state-of-the-art SEARLE with a large gap (8.07\% and 9.81\% in the average R@10 and R@50, respectively). In all categories, we observe improvements ranging from 7.57\% to 10.25\% compared to SEARLE.
In particular, we find that on the Toptee sub-dataset, our method achieved the largest 10.25\% improvement with R@50 of 56.71\%. Additionally, in terms of R@10, the proposed method yielding 26.89\% R@10 outperforms the SEARLE by improving up to 8.35\% on the Dress sub-dataset.

Considering the CLIP ViT-L/14 backbone network, SEARLE-XL-OTI \cite{Baldrati_2023_ICCV} and LinCIR \cite{gu2023languageonly} are the SOTA methods. We observe that LinCIR does not perform better than SEARLE-XL-OTI on all metrics, even though it outperforms SEARLE-XL. The proposed method with CLIP L/14 has arrived at 36.37\% Recall@10 and 58.78\% Recall@50. This shows that our method has a significant boost over the SOTA models SEARLE-XL-OTI and LinCIR. 
Moreover, the proposed method shows a significant performance boost when using the CLIP L/14 backbone compared to the CLIP B/32 backbone. This result highlights the scalability of our approach when applied to larger and more complex vision-language backbones.
Specifically, our model achieves a superior improvement of $+8.76$\% and $+10.88$\% compared to SEARLE-XL-OTI in average Recall@10 and Recall@50, respectively. In all categories, we observe improvements ranging from 7.88\% to 12.46\% compared to SEARLE-XL-OTI. The improvements are particularly notable in the Toptee category, where our method yields the largest gains. Specifically, in the Toptee category, our method with R@50 of 56.71\%, achieves the largest improvement of 12.46\%. 
Moreover, in the Toptee category, our method yields the largest improvement, with R@10 of 40.65\%, representing a gain of 9.75\%. Furthermore, our method significantly outperforms the concurrent work LinCIR on all metrics, even though our model is trained with 13\% of the pre-training data.

The proposed method shows a significant performance boost when using the CLIP L/14 backbone compared to the CLIP B/32 backbone. Specifically, the average Recall@10 improves from 30.96\% to 36.37\%, and the average Recall@50 improves from 52.34\% to 58.78\%. The performance gains are consistent across all categories (Shirt, Dress, Toptee) when considering the larger CLIP L/14 model. The greatest improvements are observed in the Toptee category, with Recall@10 and Recall@50 increasing by 7.38\% and 7.51\%, respectively. In the Dress category, while showing smaller improvements, our method still benefits from the CLIP L/14 backbone, with Recall@10 and Recall@50 increasing by 2.56\% and 4.28\%, respectively.
This result highlights the scalability of our approach when applied to larger and more complex vision-language backbones.


\begin{table*}[!htbp]
  \centering
  \caption{Quantitative Results on CIRCO Test Set. Since CIRCO Datasets Have Multiple Annotated Ground Truths, the MAP Is Used As the Metric.  Best and Second-Best Scores Are Highlighted in Bold and Underlined, Respectively. $^{\dagger}$ Indicates the Results Are Cited From~\cite{Baldrati_2023_ICCV}.  
  }
  \begin{tabular}{c|l|cccc}
  \hline
  \multicolumn{1}{c|}{} & \multicolumn{1}{c|}{} & \multicolumn{4}{c}{mAP$@K$} \\
  \cline{3-6}
  \multicolumn{1}{l|}{Backbone} & \multicolumn{1}{l|}{Method} & $K=5$ & $K=10$ & $K=25$ & $K=50$ \\ 
  \hline
  \multirow{8}{*}{CLIP B/32} & Image-only$^{\dagger}$ & 1.34 & 1.60 & 2.12 & 2.41 \\
  & Text-only$^{\dagger}$ & 2.56 & 2.67 & 2.98 & 3.18 \\
  
  & Captioning$^{\dagger}$ & 5.48 & 5.77 & 6.44 & 6.85 \\
  & PALAVRA \cite{cohen2022my}$^{\dagger}$ & 4.61 & 5.32 & 6.33 & 6.80 \\
  & {SEARLE-OTI}~\cite{Baldrati_2023_ICCV}$^{\dagger}$ & {7.14} & {7.83} & {8.99} & {9.60} \\
   & {SEARLE}~\cite{Baldrati_2023_ICCV}$^{\dagger}$ & \underline{{9.35}} & \underline{{9.94}} & \underline{{11.13}} & \underline{{11.84}} \\ 
   & Baseline$^{\dagger}$ & 2.65 & 3.25 & 4.14 & 4.54 \\
   & Ours & \textbf{10.23} & \textbf{10.58} & \textbf{11.52} & \textbf{12.12} \\
   \hline
   
  \multirow{6}{*}{CLIP L/14} & Pic2Word$^{\dagger}$~\cite{saito2023pic2word} & 8.72 & 9.51& 10.64&11.29 \\
  & {SEARLE-XL-OTI}$^{\dagger}$~\cite{Baldrati_2023_ICCV} & {10.18} & {11.03} & {12.72} & {13.67} \\
  & {SEARLE-XL}$^{\dagger}$~\cite{Baldrati_2023_ICCV} & \color{gray}{11.68}  & \color{gray}{12.73} & \color{gray}{14.33} & \color{gray}{15.12} \\
  & LinCIR \cite{gu2023languageonly} & \underline{{12.59}} & \underline{{13.58}} & \underline{{15.00}} & \underline{{15.85}} \\
  & Baseline & 4.02 & 4.93 &  6.13 & 6.78 \\
  & Ours & \textbf{13.50} & \textbf{14.20} & \textbf{15.66} & \textbf{16.39} \\
  \hline
  \end{tabular}
  \label{tab:circo_test}
\end{table*}

\subsubsection{Quantitative Results on CIRR}

As shown in Table \ref{tab:cirr_test}, the results are reported on the standard test set\footnote{\url{https://cirr.cecs.anu.edu.au/test_process/}}. 
The results demonstrate that our method consistently outperforms other methods across various Recall@K and Recall$_{\text{Subset}}@K$ metrics with both backbones (CLIP B/32 and CLIP L/14).
We emphasize that there is a significant drawback to CIRR:  captions are usually sufficient for image retrieval, and the reference images may not provide useful information. 
This problem becomes more exacerbated when considering a small subset of images.
This observation has been previously discussed in studies \cite{Baldrati_2023_ICCV, saito2023pic2word, gu2023languageonly}. In particular, the Text-only baseline attains the highest performance on the Recall$_\text{Subset}$, surpassing both the Image-only and Image + Text (Baseline) methods across all metrics. 
This suggests that visual information not only provides little benefit but may even have a detrimental effect.


With the CLIP ViT-B/32 backbone network, SEARLE \cite{Baldrati_2023_ICCV} emerges as the state-of-the-art method. Our approach surpasses SEARLE on all metrics.
For the Recall@K metrics, our method achieves the highest scores in Recall@1 (24.31\%), Recall@5 (53.76\%), Recall@10 (67.78\%), and Recall@50 (90.07\%). These results indicate an improvement over the previous best-performing methods, SEARLE and SEARLE-OTI, which achieve Recall@1 scores of 24.00\% and 24.27\%, respectively, and Recall@10 scores of 66.80\% and 66.10\%, respectively. In terms of Recall$_{\text{Subset}}@K$ metrics, our method also performs exceptionally well, achieving the second-highest scores in Recall$_{\text{Subset}}@1$ (55.66\%), Recall$_{\text{Subset}}@2$ (77.57\%), and Recall$_{\text{Subset}}@3$ (89.59\%).

When using the CLIP L/14 backbone, we compare the performance of our proposed method with two state-of-the-art methods: LinCIR \cite{gu2023languageonly} and SEARLE-XL \cite{Baldrati_2023_ICCV}. The proposed method consistently outperforms previous SOTA methods across most Recall@K and Recall$_{\text{Subset}}@K$ metrics. Compared to SEARLE-XL, our method achieves the largest improvement of 4.34\% on Recall@5. In terms of Recall$_{\text{Subset}}@K$ metrics, the largest improvements are observed in Recall$_{\text{Subset}}@2$, where our method outperforms SEARLE-XL by 2.51\%. 
Compared to LinCIR, our method improves 1.11\%, 3.57\%, and 2.62\% in terms of Recall@1, 5, and 10, respectively.
Although LinCIR slightly outperforms our method in Recall$_{\text{Subset}}@1$, the overall performance of our method remains superior. 
In summary, we follow the previous proposal of Gu et al. \cite{gu2023languageonly} and focus on the full Recall@10 scores on the CIRR dataset rather than other metrics. Our method achieves a Recall@10 score of 69.30\%, which is 3.01\% higher than SEARLE-XL and 2.62\% higher than LinCIR.

When comparing the performance of the proposed method using different backbones, CLIP B/32 and CLIP L/14, the gain by SEARLE-XL using a larger backbone on the CIRR dataset is marginal. In some cases, it may even result in a slight decrease in performance metrics. This suggests that the SEARLE method does not benefit from model enlargement.
In contrast, we observe significant improvements with the CLIP L/14 backbone network across various metrics. For instance, our method with CLIP L/14 achieves improvements in Recall@1, 5, and 10 by 1.84\%, 3.06\%, and 1.52\% respectively, compared to CLIP B/32.
This highlights our method's advantages at harnessing the potent capabilities of larger models. Hence, our method using the CLIP ViT-L/14 backbone surpasses SEARLE-XL and LinCIR to achieve SOTA performance.


\subsubsection{Quantitative Results on CIRCO}

Quantitative experimental results on the CIRCO test set\footnote{\url{https://circo.micc.unifi.it/evaluation}} are presented in Table \ref{tab:circo_test}. Overall, our proposed method consistently outperforms other comparative methods \cite{Baldrati_2023_ICCV, gu2023languageonly} across all mAP$@K$ values with both backbones (CLIP B/32 and CLIP L/14). 
Specifically, with the CLIP B/32 backbone, our method achieves the highest mAP@$K$ scores, with 10.23, 10.58, 11.52, and 12.12 at K=5, K=10, K=25, and K=50, respectively. 
Our method achieves relative improvements over SEARLE of approximately 9.41\%, 6.44\%, 3.51\%, and 2.37\% for mAP@$K$ scores at K=5, K=10, K=25, and K=50, respectively.
Additionally, it is noteworthy that our method using the CLIP B/32 backbone outperforms the Pic2Word \cite{saito2023pic2word} method with the larger CLIP L/14 backbone.
Similarly, the mAP@$K$ scores of our method with CLIP L/14 at K=5, K=10, K=25, and K=50 are 13.50, 14.20, 15.66, and 16.39, respectively, which are significantly higher than the previous SOTA method LinCIR \cite{gu2023languageonly}. Concretely, compared to LinCIR, our method achieves about 7.22\%, 4.56\%, 4.40\%, and 3.41\% relative improvement in mAP@K scores at K=5, K=10, K=25 and K=50, respectively. Moreover, we point out that it is not fair to compare our method with SEARLE-XL. This is because SEARLE-XL is employed to select and construct the CIRCO dataset, as stated in \cite{Baldrati_2023_ICCV}. Still, compared to the SEARLE-XL, our method relatively improves mAP@K scores by approximately 15.58\%, 11.54\%, 9.28\%, and 8.40\% at K=5, K=10, K=25, and K=50, respectively.
In summary, for both backbones, the performance gap between our method and other methods is relatively consistent across different mAP@$K$ metrics. 
This stable improvement shows that the proposed method does not depend on specific conditions or scenarios, but rather provides a broad-based improvement in retrieval performance.


\begin{table*}[!ht]
\centering
\caption{Quantitative Results on GeneCIS With CLIP L/14 Backbone. $^{\dagger}$ Indicates the Results Are Cited From Concurrent Work~\cite{gu2023languageonly}. Best and Second-Best Scores Are Highlighted in Bold and Underlined, Respectively.}
\begin{tabular}{l|ccc|ccc|ccc|ccc|ccc}
\hline

\multicolumn{1}{c|}{} & \multicolumn{3}{c|}{Focus Attribute} & \multicolumn{3}{c|}{Change Attribute} & \multicolumn{3}{c|}{Focus Object} & \multicolumn{3}{c|}{Change Object} & \multicolumn{3}{c}{Average}\\
\cline{2-4}
  \cline{5-7}
  \cline{8-10}
  \cline{11-13}
  \cline{14-16}
\multicolumn{1}{l|}{Method} & R@1        & R@2        & R@3       & R@1        & R@2        & R@3        & R@1       & R@2       & R@3      & R@1       & R@2       & R@3    &  R@1       & R@2       & R@3  \\
\hline
Pic2Word$^{\dagger}$ \cite{saito2023pic2word} & 15.65      & 28.16      & 38.65     & 13.87      & 24.67      & 33.05      & \underline{8.42}      & \underline{18.01}     & {25.77}    & 6.68      & 15.05     & 24.03  & 11.16 & 21.47 & 30.38   \\

SEARLE$^{\dagger}$ \cite{Baldrati_2023_ICCV} & \underline{17.00}      & 29.65      & {40.70}     & \textbf{16.38}     & 25.28      & 34.14      & 7.76      & 16.68     & 25.31    & \underline{7.91}      & \underline{16.84}     & \underline{25.05}  & \underline{12.26} &  22.11 & 31.30   \\

LinCIR$^{\dagger}$  \cite{gu2023languageonly}     & 16.90      & \underline{29.95}   & \underline{41.45}     & \underline{16.19}      & \textbf{27.98}      & \textbf{36.84}      & 8.27      & 17.40     & \underline{26.22}    & {7.40}      & {15.71}     & {25.00} & 12.19 & \underline{22.76} & \underline{32.38}    \\
Ours & \textbf{20.85} & \textbf{33.40} & \textbf{43.15} & 14.63 & \underline{26.14} & \underline{35.46} & \textbf{12.55} & \textbf{21.07} & \textbf{30.77} & \textbf{11.48} & \textbf{21.68} & \textbf{32.50} & \textbf{{14.88}} & \textbf{{25.57}} & \textbf{{35.47}} \\
\hline
\end{tabular}

\label{tab:clip_genecis}
\end{table*}
\begin{table*}[!ht]
    \centering
\caption{Quantitative Average Results on GeneCIS With CLIP L/14 Backbone. $^{\dagger}$ Indicates the Results Are Cited From Concurrent Work~\cite{gu2023languageonly}. Best and Second-Best Scores Are Highlighted in Bold and Underlined, Respectively.}
\begin{tabular}{l|ccc|ccc|ccc|ccc}
\hline

\multicolumn{1}{c|}{} & \multicolumn{3}{c|}{``Attribute'' Avg} & \multicolumn{3}{c|}{``Object'' Avg} & \multicolumn{3}{c|}{``Focus'' Avg} & \multicolumn{3}{c}{``Change'' Avg} \\
\cline{2-4}
  \cline{5-7}
  \cline{8-10}
  \cline{11-13}
\multicolumn{1}{l|}{Method} & R@1        & R@2        & R@3       & R@1        & R@2        & R@3        & R@1       & R@2       & R@3      & R@1       & R@2       & R@3  \\
\hline
Pic2Word$^{\dagger}$ \cite{saito2023pic2word} & 14.76 & 26.42 &35.85& 7.55& 16.53& 24.90& 12.04& 23.09& 32.21& 10.28 &19.86 &28.54   \\

SEARLE$^{\dagger}$ \cite{Baldrati_2023_ICCV} & \underline{16.69} & 27.47 & 37.42 &7.84& \underline{16.76} & 25.18 &12.38 &23.17& 33.01 & \underline{12.15} &21.06 &29.60   \\

LinCIR$^{\dagger}$  \cite{gu2023languageonly}     & 16.55 & \underline{28.97} & \underline{39.15} & \underline{7.84} & 16.56& \underline{25.61} & \underline{12.59} & \underline{23.68} & \underline{33.84} & 11.80 & \underline{21.85} & \underline{30.92}    \\
Ours & \textbf{17.74} & \textbf{29.77} & \textbf{39.31} & \textbf{12.02} & \textbf{21.38} & \textbf{31.63} & \textbf{16.70} & \textbf{27.24} & \textbf{36.96} & \textbf{13.06} & \textbf{23.91} & \textbf{33.98}  \\

\hline
\end{tabular}

\label{tab:clip_genecis_avg}
\end{table*}

\subsubsection{Quantitative Results on GeneCIS}
As listed in Table \ref{tab:clip_genecis}, our method significantly outperforms other methods in the average R@K results for four different subtasks. Specifically, compared to previous SOTA method LinCIR \cite{gu2023languageonly}, our method improves R@1, R@2, and R@3 by 2.69\%, 2.81\%, and 3.09\% respectively.

We further analyse the average scores for the four tasks: ``Focus'', ``Change'', ``Attribute'' and ``Object'', as shown in Table \ref{tab:clip_genecis_avg}. Overall, our method achieves the best results on all evaluation metrics compared to other methods, especially for ``Focus'' tasks and “Object” tasks. 
This validates the effectiveness of our method. We observe that our approach achieves the greatest improvement in the ``Object'' tasks. Specifically, our method achieves an improvement of 4.18\%, 4.82\%, and 6.02\% compared to LinCIR on R@1, R@2, and R@3, respectively.
Notably, we observe a significant performance gap between object and attribute subtasks for the textual inversion approach \cite{Baldrati_2023_ICCV, saito2023pic2word, gu2023languageonly}. The performance degradation on the ``Object'' tasks indicate that the textual inversion approach lacks understanding of the objects in the image and thus struggles with the object categories. Our method significantly outperforms previous methods on the ``Focus Object'' and ``Change Object'' tasks.  
The good performance across all sub-tasks indicates that our method has a strong ability to comprehend conditional similarity in a generic manner.


\subsection{Analysis}
\begin{table}[!ht]
\centering
\caption{Quantitative Results of Combiner With Frozen CLIP L/14 Backbone. The Best Scores Are Highlighted in Bold. $^{\dagger}$ Denotes the Results Are Cited From Concurrent Work~\cite{gu2023languageonly}.}
\resizebox{1\linewidth}{!}{
\begin{tabular}{lccccc}
\toprule
       & FashionIQ & CIRR  & CIRCO & GeneCIS \\
Method & R@10      & R@10  & mAP@5 & R@3     & Average\\
\midrule
Pic2Word$^{\dagger}$ \cite{saito2023pic2word} &  24.70 & 65.30 & 8.72 & 30.38 & 32.28 \\
SEARLE-XL$^{\dagger}$ \cite{Baldrati_2023_ICCV} & 25.56 & 66.29 & 11.68 & 31.30 & 33.71 \\
LinCIR$^{\dagger}$ \cite{gu2023languageonly} & 26.28  & \textbf{66.68} & \textbf{12.59} & 32.38  & 34.48 \\
Ours   & \textbf{31.52} & {64.29} & {10.03}  &  \textbf{34.66} & \textbf{35.77}\\
\bottomrule
\end{tabular}
}

\label{tab:combiner_all_res}
\end{table}

\subsubsection{Quantitative Results on Frozen VLM with Additional Network}

Our approach can be used in different cases: 1) fine-tuning the pre-trained VLM backbone, and 2) freezing the pre-training model with an additional network similar to the previous work~\cite{saito2023pic2word,gu2023languageonly}. In this set of experiments, to demonstrate the compatibility of our approach with the existing CIR methods, we evaluate our method in the case of a fixed pre-trained backbone. 

Recall that previous works \cite{Baldrati_2023_ICCV, saito2023pic2word, cohen2022my, gu2023languageonly} trained an additional text inversion. Different from them, our method does not need the text inversion. Fortunately, our work simulates the supervised triplets by masking image-text pairs. 
Thus our approach can be easily extended to existing supervised post-fusion architecture, such as Combiner \cite{baldrati2022effective}. Hence, we freeze the pre-trained CLIP L/14 backbone and train only one Combiner following \cite{baldrati2022effective}. Specifically, we first extract image and text features using frozen image and text encoders, respectively. Then, we directly use the Combiner to compose the reference image feature with the text feature. And the combined feature $f_r = Combiner(f_r^I, f_r^T)$ is obtained. 
Combiner is a lightweight network. It effectively combines the reference image and input text features to output combined features that are as close as possible to the target image. Details of the Combiner architecture can be found in \cite{baldrati2022effective}.

Table~\ref{tab:combiner_all_res} shows the comparison results of our method with the SOTA  methods~\cite{gu2023languageonly} on the four datasets. We observe that our method outperforms these methods based on text inversion in terms of average performance. Specifically, our method achieves an average score of 35.77\%, which represents an improvement of 1.29\% over LinCIR’s score of 34.48\%.
Overall, our method performs well within the context of a fixed pre-trained backbone, which demonstrates the compatibility and effectiveness of the proposed method.

\begin{table}[!ht]
\centering
 \caption{Quantitative Results on Flickr30k Dataset With CLIP L/14 Backbone. Best Scores Are Highlighted in Bold. $^{\dagger}$ Denotes the Results Are Cited From Concurrent Work~\cite{gu2023languageonly}}
\resizebox{1\linewidth}{!}{
\begin{tabular}{lccccc}
\toprule
       & FashionIQ & CIRR  & CIRCO & GeneCIS \\
Method & R@10      & R@10  & mAP@5 & R@3     & Average\\
\midrule
Pic2Word$^{\dagger}$ \cite{saito2023pic2word} &  24.70 & 65.30 & 8.72 & 30.38 & 32.28 \\
SEARLE-XL$^{\dagger}$ \cite{Baldrati_2023_ICCV} & 25.56 & 66.29 & 11.68 & 31.30 & 33.71 \\
LinCIR$^{\dagger}$ \cite{gu2023languageonly} & 26.28  & {66.68} & \textbf{12.59} & 32.38  & 34.48 \\
Ours   & \textbf{35.05} & \textbf{67.71} & {10.75}   &  \textbf{35.46} & \textbf{37.24}\\
\bottomrule
\end{tabular}
}
\label{tab:flickr_all_res}
\end{table}

\begin{figure}[htbp]
    \centering
    \includegraphics[width=0.9\linewidth]{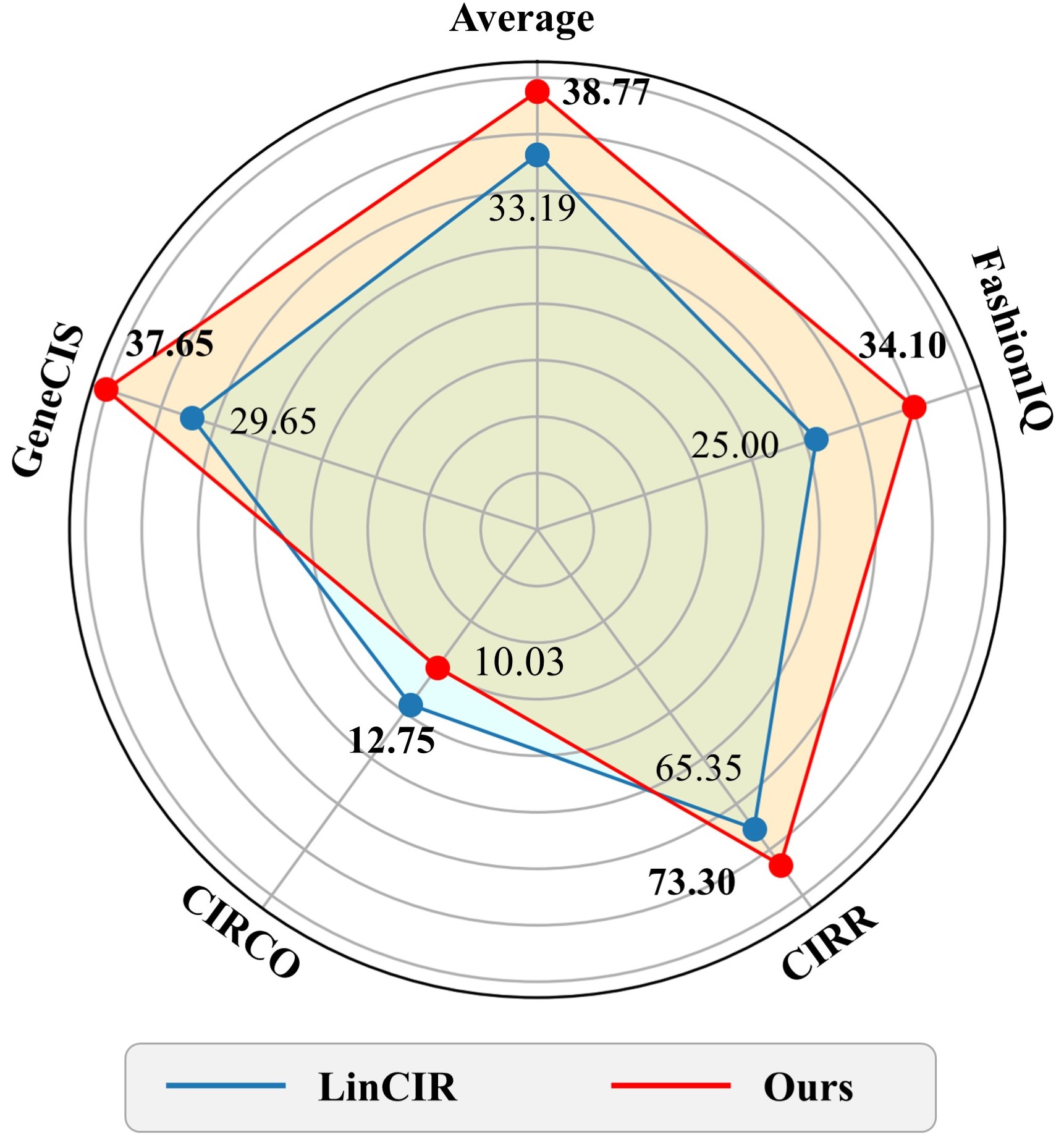}
    \caption{Quantitative results of BLIP ViT-B/16 backbone. We compare the proposed method with LinCIR \cite{gu2023languageonly} on four benchmarks. The performance on different datasets is evaluated using the average R@10 for FashionIQ and CIRR, mAP@5 for CIRCO, and average R@3 for GeneCIS.}
    \label{fig:blip_all_res}
\end{figure}

\begin{figure}[!t]
    \centering
    \subfloat[]{\includegraphics[width=0.99\linewidth]{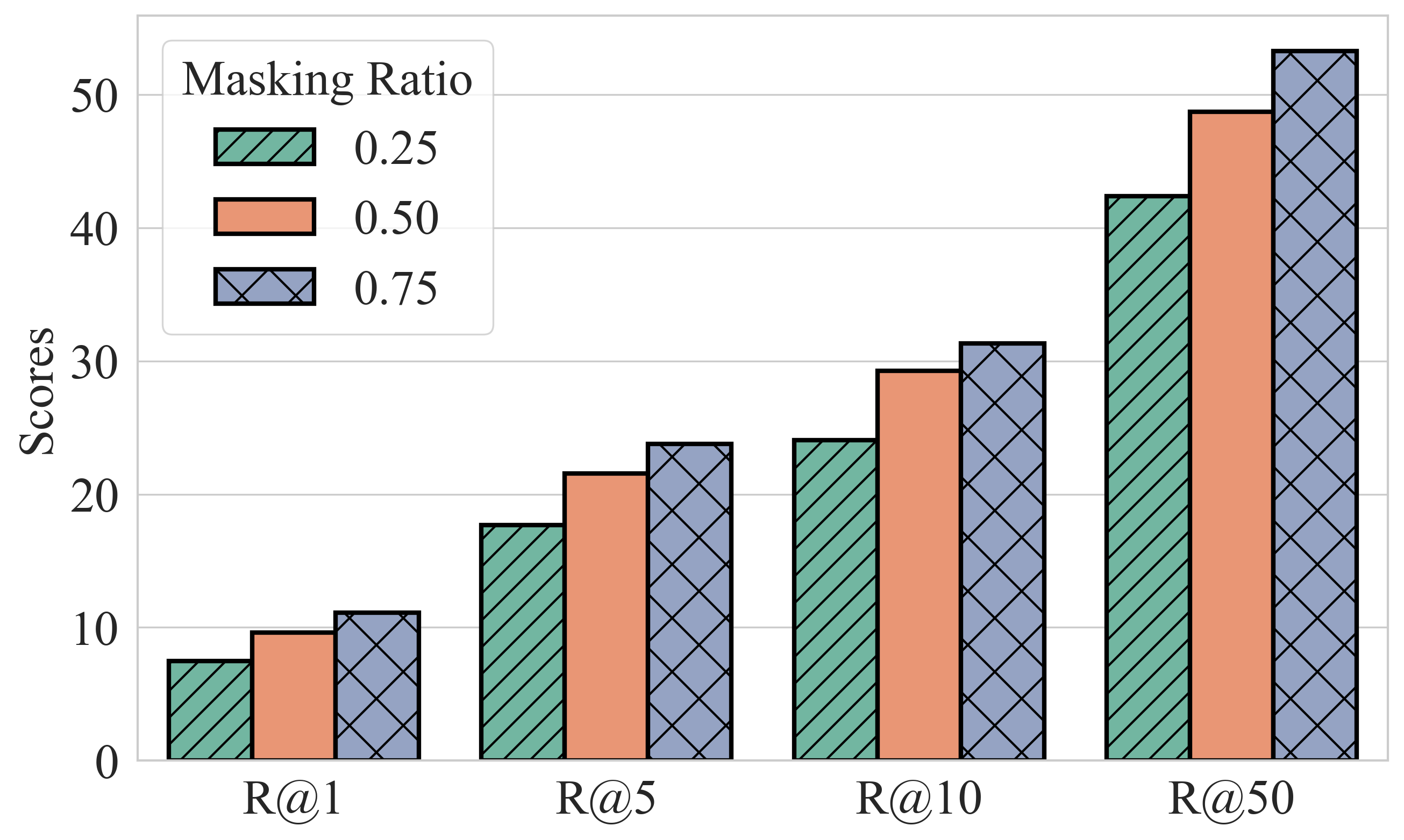}}
    \newline
    \subfloat[]{\includegraphics[width=0.99\linewidth]{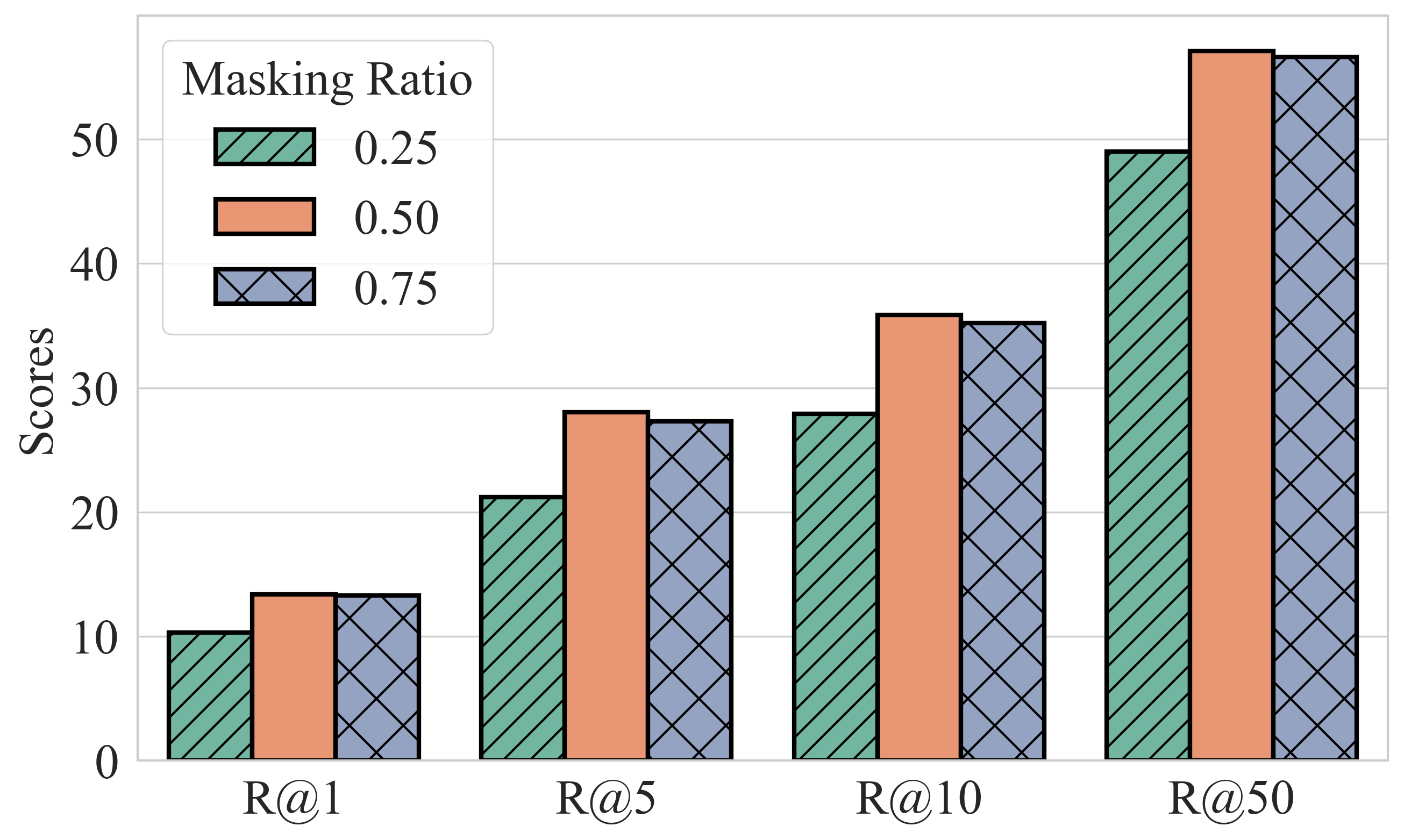}}    \caption{Influence of masking ratio $w$ on the FashionIQ dataset with different backbone: (a) CLIP B/32, (b) BLIP B/16.}
    \label{fig:ablation study}
\end{figure}

\begin{figure*}
    \centering
    \includegraphics[width=0.99\linewidth]{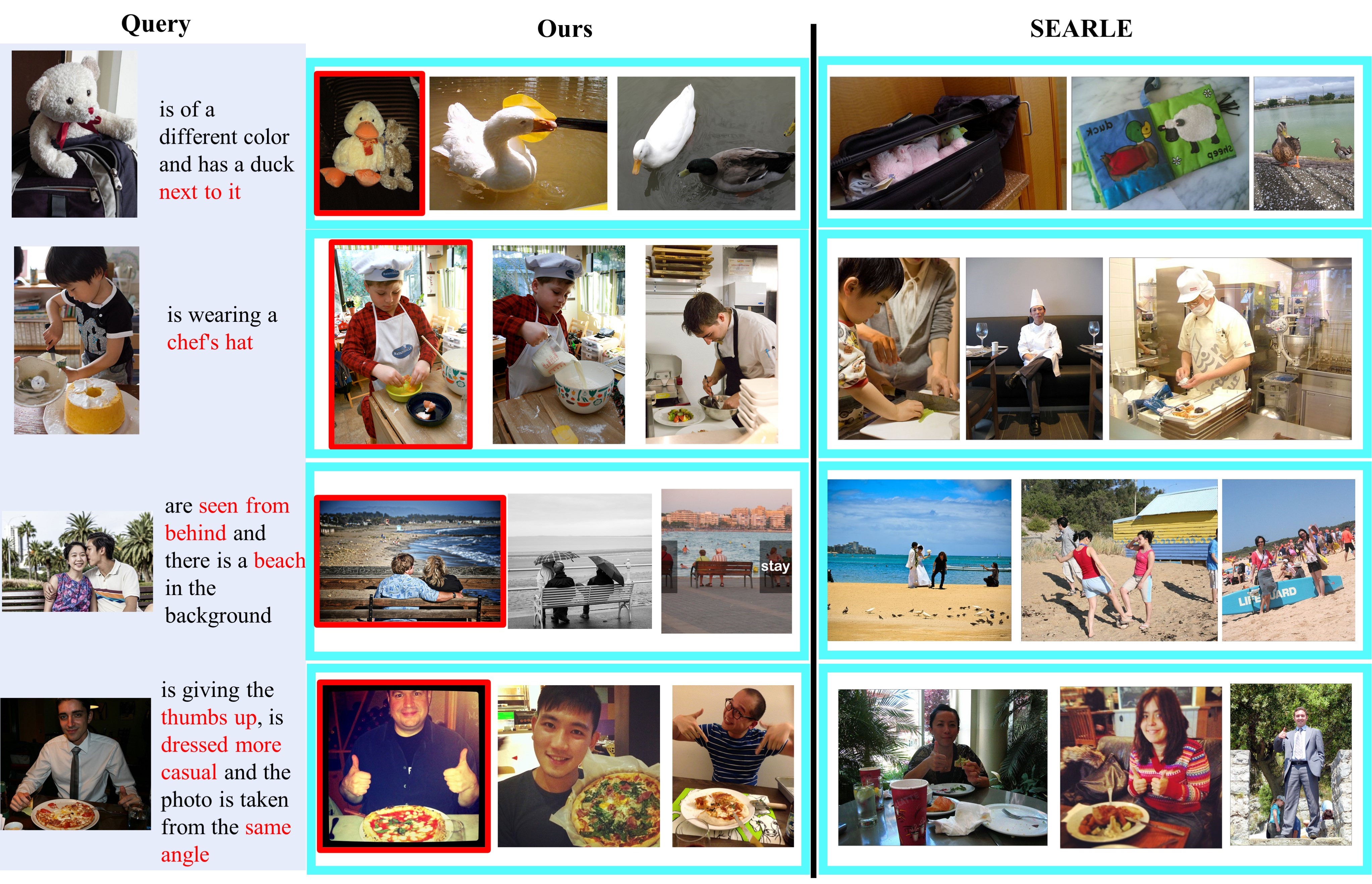}
    \caption{Top-3 examples retrieved from CIRCO validation set. 
    Ground truth retrievals are highlighted with red outline. We mainly compare the top-3 retrieved results of proposed method with the previous SOTA model SEARLE \cite{Baldrati_2023_ICCV}.}
    \label{fig:qualitative}
\end{figure*}

\subsubsection{Effects of the Small Pre-training Dataset Flickr30k}
In the above experiments, we use ImageNet and LLaVA-CC3M-Pretrain-595K as the pre-training dataset. 
In this set of experiments, we explore the impact of using smaller pre-training data for our method's performance. Specifically, we evaluate our method using the publicly available image-text dataset Flickr30k~\cite{young-etal-2014-image} as the pre-training dataset. 
Flickr30k comprises 31,783  photographs depicting everyday activities, events, and scenes. Each image is independently annotated by five distinct annotators. 
The pre-training hyper-parameters remain consistent with the above experiments.

Table~\ref{tab:flickr_all_res} shows the comparison results on the four datasets. We observe that the overall performance of our method significantly surpasses that of previous state-of-the-art methods, despite using a very small image-text dataset Flickr30k for training.
In terms of average scores, our method demonstrates superior effectiveness compared to the other approaches. Specifically, our method achieves the highest average score of 37.24\%, outperforming previous SOTA LinCIR\cite{gu2023languageonly}, which represents an improvement of 2.76\%. 
To be more precise, our method shows significant improvements on the FashionIQ, CIRR, and GeneCIS datasets. On the FashionIQ and GeneCIS dataset, our method significantly improves over LinCIR by 8.77\% R@10 and 3.08\% R@3, respectively. Additionally, on the CIRR dataset, our method slightly surpasses LinCIR by 1.03\% R@10. These results demonstrate the superiority of our method.

\subsubsection{Quantitative Results on BLIP}

In this set of experiments, we clarify that our approach can be simply extended to other VLM backbones, such as BLIP. We choose BLIP ViT-B/16 as the backbone. The mask ratio of pre-trained BLIP model is set to 50\%. We follow the same hyper-parameter settings as the CLIP backbone to utilize the AdamW optimizer and weight decay. We use ImageNet and LLaVA-CC3M-Pretrain-595K to pre-train the BLIP ViT-B/16 network.

As shown in Figure \ref{fig:blip_all_res}, we follow LinCIR \cite{gu2023languageonly} to estimate the overall performance of the BLIP backbone. 
We evaluate the performance on different datasets using the metrics average R@10 for FashionIQ and CIRR, mAP@5 for CIRCO, and average R@3 for GeneCIS. 
Our method achieves scores of 34.10\% R@10, 73.30\% R@10, 10.03\% mAP@5, and 37.65\% R@3 on FashionIQ, CIRR, CIRCO, and GeneCIS, respectively. 
Overall, our method based on the BLIP backbone outperforms the previous state-of-the-art methods LinCIR \cite{gu2023languageonly}. In terms of the average metric, our method achieves a score of 38.77\%. This represents an improvement of 5.58\% over LinCIR.
Moreover, our method demonstrates remarkable improvements on the FashionIQ, CIRR, and GeneCIS datasets, while maintaining competitive results on the CIRCO dataset.
Specifically, compared to LinCIR, our method achieves absolute boosts of 9.10\%, 7.95\%, and 8.00\% on FashionIQ, CIRR, and GeneCIS, respectively.
Extended to the BLIP backbone network, our approach also achieves state-of-the-art performance. This indicates that our approach is not only effective but also versatile, maintaining high performance across different backbone architectures and datasets.




\subsubsection{Effects of the Masking Ratio}
  
In this set of experiments, we explore the effects of different masking ratios.
Figure \ref{fig:ablation study} shows the ablation experiments conducted on our method with varying mask ratios. 
For these experiments on CLIP and BLIP, we choose three commonly used masking rates: 25\%, 50\%, and 75\%. The experimental results demonstrate that higher masking ratios (50\% or 75\%) yield superior performance, aligning with the findings in \cite{li2023scaling, he2022masked}. To be specific, CLIP ViT-B/32 with a 75\% mask rate exhibits superior performance on the FashionIQ validation set. Conversely, when the mask rate is set at 50\%, BLIP ViT-B/16 outperforms. 

\subsubsection{Qualitative Visualization}

We show the top-3 retrieval results of the proposed method and the previous method SEARLE \cite{Baldrati_2023_ICCV} on CIRCO, as shown in Figure \ref{fig:qualitative}. Through qualitative visualization, we observe that SEARLE reduces combined image retrieval to text-to-image retrieval by textual inversion and ignores the information brought by the image itself. Specifically, at the top of Figure \ref{fig:qualitative}, SEARLE is unable to understand that the ``it'' in ``next to it'' should refer to the bear of the reference image. In essence, SEARLE fails to maintain the partial feature consistency of the reference image and the retrieved image. 
For example, in the middle of Figure \ref{fig:qualitative}, the features of the query image contain two people sitting on a chair. And the text only modifies the viewpoint and background. However, we discover that the retrieved image of SEARLE does not contain the semantic features ``two people sitting on a chair''.
Similarly, at the bottom of Figure \ref{fig:qualitative}, the ``pizza'' of the reference image is missing in the retrieved image. This indicates that the text does not specify the modification of this part of the features, but the text-to-image retrieval method tends to ignore the visual information of the reference image. In comparison, our model diligently captures and preserves more comprehensive visual information. This reflects that our model can perform combined image retrieval more accurately and provide a superior user experience.

\section{Conclusion}

In this study, we strived to address the challenging task of zero-shot composed image retrieval (ZS-CIR) and reduce the reliance of retrieval models on costly manually labeled CIR data. We introduced a novel self-supervised pre-training method to intuitively narrow the gap between pre-trained models and ZS-CIR tasks. We employed a clever masking technique to generate a triplet that closely approximates the manually annotated CIR data. Furthermore, we utilized a high mask ratio during training to encourage the model to compose both image and text semantics. Experimental results on four prominent CIR benchmark datasets (FashionIQ, CIRR, GeneCIS, and CIRCO) demonstrated that our proposed method surpasses the baseline and previous competitive approaches, establishing a new state-of-the-art method.

\bibliographystyle{IEEEtran}
\bibliography{IEEEabrv, main_abrv}

\begin{thebibliography}{10}
\providecommand{\url}[1]{#1}
\csname url@samestyle\endcsname
\providecommand{\newblock}{\relax}
\providecommand{\bibinfo}[2]{#2}
\providecommand{\BIBentrySTDinterwordspacing}{\spaceskip=0pt\relax}
\providecommand{\BIBentryALTinterwordstretchfactor}{4}
\providecommand{\BIBentryALTinterwordspacing}{\spaceskip=\fontdimen2\font plus
\BIBentryALTinterwordstretchfactor\fontdimen3\font minus \fontdimen4\font\relax}
\providecommand{\BIBforeignlanguage}[2]{{%
\expandafter\ifx\csname l@#1\endcsname\relax
\typeout{** WARNING: IEEEtran.bst: No hyphenation pattern has been}%
\typeout{** loaded for the language `#1'. Using the pattern for}%
\typeout{** the default language instead.}%
\else
\language=\csname l@#1\endcsname
\fi
#2}}
\providecommand{\BIBdecl}{\relax}
\BIBdecl

\bibitem{baldrati2022effective}
A.~Baldrati, M.~Bertini, T.~Uricchio, and A.~Del~Bimbo, ``Effective conditioned and composed image retrieval combining clip-based features,'' in \emph{Proc. {IEEE} Conf. Comput. Vis. Pattern Recog.}, 2022, pp. 21\,466--21\,474.

\bibitem{liu2021image}
Z.~Liu, C.~Rodriguez-Opazo, D.~Teney, and S.~Gould, ``Image retrieval on real-life images with pre-trained vision-and-language models,'' in \emph{Proc. {IEEE} Conf. Comput. Vis. Pattern Recog.}, 2021, pp. 2125--2134.

\bibitem{lee2021cosmo}
S.~Lee, D.~Kim, and B.~Han, ``Cosmo: Content-style modulation for image retrieval with text feedback,'' in \emph{Proc. {IEEE} Conf. Comput. Vis. Pattern Recog.}, 2021, pp. 802--812.

\bibitem{xu2024csvt}
Y.~Xu, J.~Wei, Y.~Bin, Y.~Yang, Z.~Ma, and H.~T. Shen, ``Set of diverse queries with uncertainty regularization for composed image retrieval,'' \emph{{IEEE} Trans. Circuits Syst. Video Technol.}, pp. 1--1, 2024, doi: {10.1109/TCSVT.2024.3401006}.

\bibitem{vo2019composing}
N.~Vo, L.~Jiang, C.~Sun, K.~Murphy, L.-J. Li, L.~Fei-Fei, and J.~Hays, ``Composing text and image for image retrieval-an empirical odyssey,'' in \emph{Proc. {IEEE} Conf. Comput. Vis. Pattern Recog.}, 2019, pp. 6439--6448.

\bibitem{jandial2022sac}
S.~Jandial, P.~Badjatiya, P.~Chawla, A.~Chopra, M.~Sarkar, and B.~Krishnamurthy, ``Sac: Semantic attention composition for text-conditioned image retrieval,'' in \emph{Proc. {IEEE} Winter Conf. Appl. Comput. Vis.}, 2022, pp. 4021--4030.

\bibitem{wu2021fashion}
H.~Wu, Y.~Gao, X.~Guo, Z.~Al-Halah, S.~Rennie, K.~Grauman, and R.~Feris, ``Fashion iq: A new dataset towards retrieving images by natural language feedback,'' in \emph{Proc. {IEEE} Conf. Comput. Vis. Pattern Recog.}, 2021, pp. 11\,307--11\,317.

\bibitem{li2022blip}
J.~Li, D.~Li, C.~Xiong, and S.~Hoi, ``Blip: Bootstrapping language-image pre-training for unified vision-language understanding and generation,'' in \emph{Proc. Int. Conf. Mach. Learn.}\hskip 1em plus 0.5em minus 0.4em\relax PMLR, 2022, pp. 12\,888--12\,900.

\bibitem{radford2021clip}
A.~Radford, J.~W. Kim, C.~Hallacy, A.~Ramesh, G.~Goh, S.~Agarwal, G.~Sastry, A.~Askell, P.~Mishkin, J.~Clark \emph{et~al.}, ``Learning transferable visual models from natural language supervision,'' in \emph{Proc. Int. Conf. Mach. Learn.}\hskip 1em plus 0.5em minus 0.4em\relax PMLR, 2021, pp. 8748--8763.

\bibitem{Baldrati_2023_ICCV}
A.~Baldrati, L.~Agnolucci, M.~Bertini, and A.~Del~Bimbo, ``Zero-shot composed image retrieval with textual inversion,'' in \emph{Proc. Int. Conf. Comput. Vis.}, October 2023, pp. 15\,338--15\,347.

\bibitem{saito2023pic2word}
K.~Saito, K.~Sohn, X.~Zhang, C.-L. Li, C.-Y. Lee, K.~Saenko, and T.~Pfister, ``Pic2word: Mapping pictures to words for zero-shot composed image retrieval,'' in \emph{Proc. {IEEE} Conf. Comput. Vis. Pattern Recog.}, 2023, pp. 19\,305--19\,314.

\bibitem{baldrati2022conditioned}
A.~Baldrati, M.~Bertini, T.~Uricchio, and A.~Del~Bimbo, ``Conditioned and composed image retrieval combining and partially fine-tuning clip-based features,'' in \emph{Proc. {IEEE} Conf. Comput. Vis. Pattern Recog.}, 2022, pp. 4959--4968.

\bibitem{liu2023bi}
Z.~Liu, W.~Sun, Y.~Hong, D.~Teney, and S.~Gould, ``Bi-directional training for composed image retrieval via text prompt learning,'' \emph{arXiv preprint arXiv:2303.16604}, 2023.

\bibitem{karthik2024visionbylanguage}
\BIBentryALTinterwordspacing
S.~Karthik, K.~Roth, M.~Mancini, and Z.~Akata, ``Vision-by-language for training-free compositional image retrieval,'' in \emph{Proc. Int. Conf. Learn. Represent.}, 2024. [Online]. Available: \url{https://openreview.net/forum?id=EDPxCjXzSb}
\BIBentrySTDinterwordspacing

\bibitem{liimproving}
W.~Li, H.~Fan, Y.~Wong, Y.~Yang, and M.~Kankanhalli, ``Improving context understanding in multimodal large language models via multimodal composition learning,'' in \emph{Proc. Int. Conf. Mach. Learn.}\hskip 1em plus 0.5em minus 0.4em\relax PMLR, 2024.

\bibitem{cohen2022my}
N.~Cohen, R.~Gal, E.~A. Meirom, G.~Chechik, and Y.~Atzmon, ``{``This is my unicorn, Fluffy": Personalizing frozen vision-language representations},'' in \emph{Proc. Eur. Conf. Comput. Vis.}\hskip 1em plus 0.5em minus 0.4em\relax Springer, 2022, pp. 558--577.

\bibitem{gu2023languageonly}
G.~Gu, S.~Chun, W.~Kim, Y.~Kang, and S.~Yun, ``Language-only training of zero-shot composed image retrieval,'' in \emph{Proc. {IEEE} Conf. Comput. Vis. Pattern Recog.}, June 2024, pp. 13\,225--13\,234.

\bibitem{meng2024csvt}
M.~Meng, J.~Sun, J.~Liu, J.~Yu, and J.~Wu, ``Semantic disentanglement adversarial hashing for cross-modal retrieval,'' \emph{{IEEE} Trans. Circuits Syst. Video Technol.}, vol.~34, no.~3, pp. 1914--1926, 2024.

\bibitem{wei2023csvt}
J.~Wei, Y.~Yang, X.~Xu, J.~Song, G.~Wang, and H.~T. Shen, ``Less is better: Exponential loss for cross-modal matching,'' \emph{{IEEE} Trans. Circuits Syst. Video Technol.}, vol.~33, no.~9, pp. 5271--5280, 2023.

\bibitem{li2021align}
J.~Li, R.~Selvaraju, A.~Gotmare, S.~Joty, C.~Xiong, and S.~C.~H. Hoi, ``Align before fuse: Vision and language representation learning with momentum distillation,'' in \emph{Proc. Adv. Neural Inform. Process. Syst.}, vol.~34, 2021, pp. 9694--9705.

\bibitem{wu2024tmmResolving}
S.~Wu, G.~Zhao, and X.~Qian, ``Resolving zero-shot and fact-based visual question answering via enhanced fact retrieval,'' \emph{{IEEE} Trans. Multimedia}, vol.~26, pp. 1790--1800, 2024.

\bibitem{chen2024csvt_vqa}
Z.~Chen, L.~Wang, P.~Wang, and P.~Gao, ``Question-aware global-local video understanding network for audio-visual question answering,'' \emph{{IEEE} Trans. Circuits Syst. Video Technol.}, vol.~34, no.~5, pp. 4109--4119, 2024.

\bibitem{hu2022scaling}
X.~Hu, Z.~Gan, J.~Wang, Z.~Yang, Z.~Liu, Y.~Lu, and L.~Wang, ``Scaling up vision-language pre-training for image captioning,'' in \emph{Proc. {IEEE} Conf. Comput. Vis. Pattern Recog.}, 2022, pp. 17\,980--17\,989.

\bibitem{yu2020csvt_ic}
J.~Yu, J.~Li, Z.~Yu, and Q.~Huang, ``Multimodal transformer with multi-view visual representation for image captioning,'' \emph{{IEEE} Trans. Circuits Syst. Video Technol.}, vol.~30, no.~12, pp. 4467--4480, 2020.

\bibitem{rombach2022high}
R.~Rombach, A.~Blattmann, D.~Lorenz, P.~Esser, and B.~Ommer, ``High-resolution image synthesis with latent diffusion models,'' in \emph{Proc. {IEEE} Conf. Comput. Vis. Pattern Recog.}, 2022, pp. 10\,684--10\,695.

\bibitem{gao2024csvt_ig}
F.~Gao, X.~Deng, J.~Jing, X.~Zou, and M.~Xu, ``Extremely low bit-rate image compression via invertible image generation,'' \emph{{IEEE} Trans. Circuits Syst. Video Technol.}, vol.~34, no.~8, pp. 6993--7004, 2024.

\bibitem{chen2022composed}
Y.~Chen, Z.~Zheng, W.~Ji, L.~Qu, and T.-S. Chua, ``Composed image retrieval with text feedback via multi-grained uncertainty regularization,'' \emph{arXiv preprint arXiv:2211.07394}, 2022.

\bibitem{chen2023ranking}
J.~Chen and H.~Lai, ``Ranking-aware uncertainty for text-guided image retrieval,'' \emph{arXiv preprint arXiv:2308.08131}, 2023.

\bibitem{liu2023zero}
Y.~Liu, J.~Yao, Y.~Zhang, Y.~Wang, and W.~Xie, ``Zero-shot composed text-image retrieval,'' \emph{arXiv preprint arXiv:2306.07272}, 2023.

\bibitem{gu2023compodiff}
G.~Gu, S.~Chun, W.~Kim, H.~Jun, Y.~Kang, and S.~Yun, ``Compodiff: Versatile composed image retrieval with latent diffusion,'' \emph{arXiv preprint arXiv:2303.11916}, 2023.

\bibitem{vaze2023genecis}
S.~Vaze, N.~Carion, and I.~Misra, ``Genecis: A benchmark for general conditional image similarity,'' in \emph{Proc. {IEEE} Conf. Comput. Vis. Pattern Recog.}, 2023, pp. 6862--6872.

\bibitem{LaSCo_2024}
M.~Levy, R.~Ben-Ari, N.~Darshan, and D.~Lischinski, ``Data roaming and quality assessment for composed image retrieval,'' in \emph{Proc. AAAI Conf. Artif. Intell.}, vol.~38, no.~4, Mar. 2024, pp. 2991--2999.

\bibitem{Jang_2024_CVPR}
Y.~K. Jang, D.~Kim, Z.~Meng, D.~Huynh, and S.-N. Lim, ``Visual delta generator with large multi-modal models for semi-supervised composed image retrieval,'' in \emph{Proc. {IEEE} Conf. Comput. Vis. Pattern Recog.}, June 2024, pp. 16\,805--16\,814.

\bibitem{brown2020language}
T.~Brown, B.~Mann, N.~Ryder, M.~Subbiah, J.~D. Kaplan, P.~Dhariwal, A.~Neelakantan, P.~Shyam, G.~Sastry, A.~Askell \emph{et~al.}, ``Language models are few-shot learners,'' in \emph{Proc. Adv. Neural Inform. Process. Syst.}, vol.~33, 2020, pp. 1877--1901.

\bibitem{CoVR_2024}
L.~Ventura, A.~Yang, C.~Schmid, and G.~Varol, ``Co{VR}: Learning composed video retrieval from web video captions,'' in \emph{Proc. AAAI Conf. Artif. Intell.}, vol.~38, no.~6, Mar. 2024, pp. 5270--5279.

\bibitem{koh2023grounding}
J.~Y. Koh, R.~Salakhutdinov, and D.~Fried, ``Grounding language models to images for multimodal inputs and outputs,'' in \emph{Proc. Int. Conf. Mach. Learn.}\hskip 1em plus 0.5em minus 0.4em\relax PMLR, 2023, pp. 17\,283--17\,300.

\bibitem{tang2023contexti2w}
Y.~Tang, J.~Yu, K.~Gai, J.~Zhuang, G.~Xiong, Y.~Hu, and Q.~Wu, ``Context-{I}2{W}: Mapping images to context-dependent words for accurate zero-shot composed image retrieval,'' in \emph{Proc. AAAI Conf. Artif. Intell.}, vol.~38, no.~6, Mar. 2024, pp. 5180--5188.

\bibitem{du2024imagesentence}
\BIBentryALTinterwordspacing
Y.~Du, M.~Wang, W.~Zhou, S.~Hui, and H.~Li, ``Image2sentence based asymmetrical zero-shot composed image retrieval,'' in \emph{Proc. Int. Conf. Learn. Represent.}, 2024. [Online]. Available: \url{https://openreview.net/forum?id=5BXAXOpaWu}
\BIBentrySTDinterwordspacing

\bibitem{kenton2019bert}
J.~D. M.-W.~C. Kenton and L.~K. Toutanova, ``Bert: Pre-training of deep bidirectional transformers for language understanding,'' in \emph{Conf. N. Am. Chapter Assoc. Comput. Linguist.}, 2019, pp. 4171--4186.

\bibitem{zhang2024tmmmask}
K.~Zhang, Y.~Yang, J.~Yu, H.~Jiang, J.~Fan, Q.~Huang, and W.~Han, ``Multi-task paired masking with alignment modeling for medical vision-language pre-training,'' \emph{{IEEE} Trans. Multimedia}, vol.~26, pp. 4706--4721, 2024.

\bibitem{dosovitskiy2020vit}
A.~Dosovitskiy, L.~Beyer, A.~Kolesnikov, D.~Weissenborn, X.~Zhai, T.~Unterthiner, M.~Dehghani, M.~Minderer, G.~Heigold, S.~Gelly \emph{et~al.}, ``An image is worth 16x16 words: Transformers for image recognition at scale,'' in \emph{Proc. Int. Conf. Learn. Represent.}, 2021.

\bibitem{wen2024image}
Z.~Wen, J.~Qian, B.~Qian, Q.~Yuan, J.~Qin, Q.~Xuan, and Y.~Yuan, ``Across images and graphs for question answering,'' in \emph{Proc. Int. Conf. Data. Eng.}, 2024, pp. 1366--1379.

\bibitem{fang2024mae}
Y.~Fang, J.~Xie, Y.~Zhao, L.~Chen, Y.~Gao, and K.~Zheng, ``Temporal-frequency masked autoencoders for time series anomaly detection,'' in \emph{Proc. Int. Conf. Data. Eng.}, 2024, pp. 1228--1241.

\bibitem{vaswani2017attention}
A.~Vaswani, N.~Shazeer, N.~Parmar, J.~Uszkoreit, L.~Jones, A.~N. Gomez, {\L}.~Kaiser, and I.~Polosukhin, ``Attention is all you need,'' in \emph{Proc. Adv. Neural Inform. Process. Syst.}, vol.~30, 2017.

\bibitem{chen2020generative}
M.~Chen, A.~Radford, R.~Child, J.~Wu, H.~Jun, D.~Luan, and I.~Sutskever, ``Generative pretraining from pixels,'' in \emph{Proc. Int. Conf. Mach. Learn.}\hskip 1em plus 0.5em minus 0.4em\relax PMLR, 2020, pp. 1691--1703.

\bibitem{bao2021beit}
H.~Bao, L.~Dong, S.~Piao, and F.~Wei, ``Beit: Bert pre-training of image transformers,'' in \emph{Proc. Int. Conf. Learn. Represent.}, 2021.

\bibitem{he2022masked}
K.~He, X.~Chen, S.~Xie, Y.~Li, P.~Doll{\'a}r, and R.~Girshick, ``Masked autoencoders are scalable vision learners,'' in \emph{Proc. {IEEE} Conf. Comput. Vis. Pattern Recog.}, 2022, pp. 16\,000--16\,009.

\bibitem{li2023scaling}
Y.~Li, H.~Fan, R.~Hu, C.~Feichtenhofer, and K.~He, ``Scaling language-image pre-training via masking,'' in \emph{Proc. {IEEE} Conf. Comput. Vis. Pattern Recog.}, 2023, pp. 23\,390--23\,400.

\bibitem{chen2020simple}
T.~Chen, S.~Kornblith, M.~Norouzi, and G.~Hinton, ``A simple framework for contrastive learning of visual representations,'' in \emph{Proc. Int. Conf. Mach. Learn.}\hskip 1em plus 0.5em minus 0.4em\relax PMLR, 2020, pp. 1597--1607.

\bibitem{sharma2018conceptual}
P.~Sharma, N.~Ding, S.~Goodman, and R.~Soricut, ``Conceptual captions: A cleaned, hypernymed, image alt-text dataset for automatic image captioning,'' in \emph{Proc. Annu. Meet. Assoc. Comput. Linguist.}, 2018, pp. 2556--2565.

\bibitem{liu2024visual}
H.~Liu, C.~Li, Q.~Wu, and Y.~J. Lee, ``Visual instruction tuning,'' in \emph{Proc. Adv. Neural Inform. Process. Syst.}, vol.~36.\hskip 1em plus 0.5em minus 0.4em\relax Curran Associates, Inc., 2023, pp. 34\,892--34\,916.

\bibitem{russakovsky2015imagenet}
O.~Russakovsky, J.~Deng, H.~Su, J.~Krause, S.~Satheesh, S.~Ma, Z.~Huang, A.~Karpathy, A.~Khosla, M.~Bernstein \emph{et~al.}, ``Imagenet large scale visual recognition challenge,'' \emph{Int. J. Comput. Vis.}, vol. 115, pp. 211--252, 2015.

\bibitem{loshchilov2019decoupled}
I.~Loshchilov and F.~Hutter, ``Decoupled weight decay regularization,'' in \emph{Proc. Int. Conf. Learn. Represent.}, 2019.

\bibitem{NEURIPS2019_bdbca288}
A.~Paszke, S.~Gross, F.~Massa, A.~Lerer, J.~Bradbury, G.~Chanan, T.~Killeen, Z.~Lin, N.~Gimelshein, L.~Antiga, A.~Desmaison, A.~Kopf, E.~Yang, Z.~DeVito, M.~Raison, A.~Tejani, S.~Chilamkurthy, B.~Steiner, L.~Fang, J.~Bai, and S.~Chintala, ``Pytorch: An imperative style, high-performance deep learning library,'' in \emph{Proc. Adv. Neural Inform. Process. Syst.}, vol.~32.\hskip 1em plus 0.5em minus 0.4em\relax Curran Associates, Inc., 2019, pp. 8024--8035.

\bibitem{Chen_2020_CVPR}
Y.~Chen, S.~Gong, and L.~Bazzani, ``Image search with text feedback by visiolinguistic attention learning,'' in \emph{Proc. {IEEE} Conf. Comput. Vis. Pattern Recog.}, 2020, pp. 2998--3008.

\bibitem{suhr-etal-2019-corpus}
A.~Suhr, S.~Zhou, A.~Zhang, I.~Zhang, H.~Bai, and Y.~Artzi, ``A corpus for reasoning about natural language grounded in photographs,'' in \emph{Proc. Annu. Meet. Assoc. Comput. Linguist.}, Florence, Italy, Jul. 2019, pp. 6418--6428.

\bibitem{lin2014microsoft}
T.-Y. Lin, M.~Maire, S.~Belongie, J.~Hays, P.~Perona, D.~Ramanan, P.~Doll{\'a}r, and C.~L. Zitnick, ``Microsoft coco: Common objects in context,'' in \emph{Proc. Eur. Conf. Comput. Vis.}\hskip 1em plus 0.5em minus 0.4em\relax Springer, 2014, pp. 740--755.

\bibitem{pham2021learning}
K.~Pham, K.~Kafle, Z.~Lin, Z.~Ding, S.~Cohen, Q.~Tran, and A.~Shrivastava, ``Learning to predict visual attributes in the wild,'' in \emph{Proc. {IEEE} Conf. Comput. Vis. Pattern Recog.}, 2021, pp. 13\,018--13\,028.

\bibitem{young-etal-2014-image}
\BIBentryALTinterwordspacing
P.~Young, A.~Lai, M.~Hodosh, and J.~Hockenmaier, ``From image descriptions to visual denotations: New similarity metrics for semantic inference over event descriptions,'' \emph{Trans. Assoc. Comput. Linguist.}, vol.~2, pp. 67--78, 2014. [Online]. Available: \url{https://aclanthology.org/Q14-1006}
\BIBentrySTDinterwordspacing

\end{thebibliography}




\end{document}